\documentclass[12pt]{article}

\usepackage[utf8]{inputenc} 
\usepackage[english]{babel} 
\usepackage{mathtools}
\usepackage{xcolor}
\usepackage[export]{adjustbox} 
\usepackage{amsmath} 
\allowdisplaybreaks
\usepackage{amssymb} 
\usepackage{anysize} 
    \marginsize{2cm}{2cm}{2cm}{2cm} 
\usepackage{appendix} 
\usepackage{cancel} 
\usepackage{caption} 
    \DeclareCaptionFont{newfont}{\fontfamily{cmss}\selectfont}
\usepackage{cite} 
\usepackage{color} 
\usepackage{fancyhdr} 
\usepackage{float} 
\usepackage{graphicx} 
\usepackage[colorlinks = true, plainpages = true, linkcolor = istblue, urlcolor = istblue, citecolor = istblue, anchorcolor = istblue]{hyperref}
\usepackage{indentfirst} 
\usepackage[super]{nth} 
\usepackage{siunitx} 
\usepackage{subcaption} 
\usepackage{tcolorbox}
\usepackage{wrapfig}
\usepackage{titlesec} 
    \titleformat{\section}{\fontfamily{cmss}\selectfont\Large\bfseries}{\thesection}{1em}{}
    \titleformat{\subsection}{\fontfamily{cmss}\selectfont\large\bfseries}{\thesubsection}{1em}{}
    \titleformat{\subsubsection}{\fontfamily{cmss}\selectfont\normalsize\bfseries}{\thesubsubsection}{1em}{}
\usepackage{algpseudocode}

\def\@Aboxed#1&#2\ENDDNE{%
  \settowidth\@tempdima{$\displaystyle#1{}$}%
  \addtolength\@tempdima{\fboxsep}%
  \addtolength\@tempdima{\fboxrule}%
  \global\@tempdima=\@tempdima
  \kern\@tempdima
  &
  \kern-\@tempdima
  \fcolorbox{gray!20}{gray!20}{$\displaystyle #1#2$}
}

\usepackage{algorithm}
\usepackage{lipsum}
\usepackage{duckuments}
\newtheorem{theorem}{Theorem}[section]

\newtheorem{proposition}[theorem]{Proposition}

\newtheorem{definition}[theorem]{Definition}
\newcommand{\sen}{\operatorname{\sen}} 
\newcommand{\HRule}{\rule{\linewidth}{0.5mm}} 

\definecolor{istblue}{RGB}{3, 171, 230}
\definecolor{dkgreen}{rgb}{0,0.6,0}
\definecolor{gray}{rgb}{0.5,0.5,0.5}

\begin{document}

\begin{center}
    \mbox{}\\[2.0cm]
    \textsc{\LARGE Generative Diffusion Modeling}\\[2.0cm]
    \HRule\\[0.4cm]
    {\large \bf {\fontfamily{cmss}\selectfont A Practical Handbook} \\[0.2cm]}
    \HRule\\[1.5cm]
\end{center}

\begin{flushleft}
    \textbf{\fontfamily{cmss}\selectfont Authors:}
\end{flushleft}

\begin{center}
    \begin{minipage}{0.5\textwidth}
        \begin{flushleft}
            Zihan Ding\\
            Chi Jin
        \end{flushleft}
    \end{minipage}%
    \begin{minipage}{0.5\textwidth}
        \begin{flushright}
            \href{mailto:zihand@princeton.edu}{\texttt{zihand@princeton.edu}}\\
            \href{mailto:chij@princeton.edu}{\texttt{chij@princeton.edu}}\\
        \end{flushright}
    \end{minipage}
\end{center}
    
    

\thispagestyle{empty}

\setcounter{page}{0}

\newpage


\newpage
\begin{abstract}
This handbook offers a unified perspective on diffusion models, encompassing diffusion probabilistic models, score-based generative models, consistency models, rectified flow, and related methods. By standardizing notations and aligning them with code implementations, it aims to bridge the ``paper-to-code" gap and facilitate robust implementations and fair comparisons. The content encompasses the fundamentals of diffusion models, the pre-training process, and various post-training methods. Post-training techniques include model distillation and reward-based fine-tuning. Designed as a practical guide, it emphasizes clarity and usability over theoretical depth, focusing on widely adopted approaches in generative modeling with diffusion models.

\end{abstract}

\tableofcontents 

\section{Introduction}
Diffusion models~\cite{sohl2015deep, song2019generative, ho2020denoising, song2021score} have recently revolutionized generative modeling, achieving remarkable advancements in generating images~\cite{rombach2022high, ramesh2022hierarchical}, audio~\cite{liu2023audioldm, polyak2024movie}, video~\cite{brooks2024video, polyak2024movie}, 3D content~\cite{poole2022dreamfusion, wang2024prolificdreamer}, and even 4D representations~\cite{wang2024vidu4d}. Despite this rapid progress, existing research often presents the concepts and notations of diffusion models in inconsistent ways, even when the underlying methodologies are similar. Compounding this issue, the formulations in many papers often differ from their their corresponding code implementations or lack sufficient implementation details, creating what is referred to as the ``paper-to-code" gap. These inconsistencies make it challenging for practitioners to directly implement methods from the literature, hindering robust implementations and fair cross-method comparisons.

This handbook seeks to provide a unified perspective on various models, collectively referred to here as ``diffusion models" as an umbrella term\footnote{This might be inappropriate, as some may differentiate score-based generative models from diffusion probabilistic models and instead prefer the umbrella term ``generative diffusion processes''.}. These include diffusion probabilistic models~\cite{sohl2015deep, ho2020denoising}, score-based generative models~\cite{song2021score}, consistency models~\cite{song2023consistency}, rectified flow~\cite{liu2022flow}, flow matching~\cite{lipman2022flow}, TrigFlow~\cite{lu2024simplifying}, and others. The authors have made every effort to standardize the notations across these methods and align them closely with their code implementations to minimize the paper-to-code gap. Additionally, this handbook aims to clarify the relationships among these different methods, highlighting existing transformations and unifications that connect them.

While diffusion models are theoretically grounded in fields such as non-equilibrium statistical physics, optimal transport, and score matching, this handbook intentionally avoids delving deeply into these theoretical foundations. Instead, its primary purpose is to serve as a practical guide for researchers and practitioners.

It is important to note that this handbook does not aim to provide a comprehensive overview of diffusion models for generative modeling. Rather, it focuses on a subset of the most popular and widely adopted works that are frequently applied in practice.

\newpage
\section*{Notations}
\begin{tabular}{cp{0.8\textwidth}}
  \( a, b, c \) & Vectors or scalars or constants. \\
  \( \mathbf{A} \) & A matrix. \\
  \(q(x_t|x_{t-1})\) & Single-step conditional distribution of diffusion process. \\
  \(q(x_{t-1}|x_t)\) & Single-step posterior distribution of denoising process in diffusion. \\
  \(p_\theta(x_{t-1}|x_t)\) & Neural network approximated single-step distribution of denoising process in diffusion. \\
  \( q(x) \) & Sample distribution of training dataset.\\
  \( x_0 \) & Clean sample from sample distribution, $0$ indicates timestep $t=0$ in diffusion process. \\
  \( \hat{x}_0 \) & Predicted clean sample. \\
  \( x_\theta \) & Predicted clean sample $\hat{x}_0$ with model parameterized by $\theta$. \\
  \( \mathcal{N}(0, \mathbf{I}) \) & Normal distribution.\\
  \( \varepsilon \) & Sampled Gaussian noise. \\
  \( \epsilon_\theta \) & Predicted noise with parameterized model by $\theta$. \\
\end{tabular}

The above notations are kept consistent across the entire paper.

\newpage
\section{Diffusion Model Basics}
\label{sec:diffusion_basic}
\subsection{Problem Formulation}
We consider samples $x^{(1)}, x^{(2)}, \dots x^{(n)}\sim q(x)$, and try to fit a parameterized distribution $p_\theta(x)$ from these samples. After that, new samples can be drawn $x\sim p_\theta(x)$.

The classical choice of $p_\theta(x)$ can be Gaussian distributions, Gaussian mixture models, etc.

In general, the maximum likelihood estimation $\max_\theta \sum_i^n \log p_\theta(x=x^{(i)})$ is intractable for the integration in the log-probability estimation:
\begin{align*}
\log p_\theta(x) &= \log \int p_\theta(x, z) dz\\
&= \log \int p_\theta(x|z)p(z) dz 
\end{align*}

Variational inference offers a way:
\begin{align*}
\log p_\theta(x) &= \log \int p_\theta(x, z) dz\\
&= \log \int p_\theta(x|z)p(z) dz \\
&=\log \int q_\phi(z|x) \frac{p_\theta(x|z)p(z)}{q_\phi(z|x)} dz \quad\quad \text{(with approximate posterior $q_\phi(z|x)$)}\\
&\geq \mathbb{E}_{q_\phi(z|x)} \left[ \log p_\theta(x|z) + \log p(z) - \log {q_\phi(z|x)} \right]  \quad\quad \text{(Jensen's Inequality)}\\
\end{align*}
with the evidence lower bound (ELBO) $\mathcal{L}(\phi, \theta; x) = \mathbb{E}_{q_\phi(z|x)} \left[ \log p_\theta(x|z) \right] - D_{\text{KL}}(q_\phi(z|x) \| p(z))$ as a tractable objective. With additional loss term $D_{\text{KL}}(q_\phi(z|x) \| p_\theta(z|x))$, it constitutes the variational auto-encoder (VAE)~\cite{kingma2013auto}. 

Apart from variational autoencoders (VAEs), other generative methods such as energy-based models, normalizing flows, and generative adversarial networks (GANs) offer diverse approaches to estimating the probability distributions of training data. Among these, diffusion models stand out for their exceptional tractability and flexibility.

\begin{definition}[Diffusion Model]
For the data sampled from distribution $x_0\sim q(x_0)$, the forward diffusion process corrupts the data distribution to a Gaussian distribution as following: 
\begin{align*}
x_t=\bar{a}_t x_0 + \bar{b}_t \varepsilon, \varepsilon\sim\mathcal{N}(0, \mathbf{I})
\end{align*}
with $t\in\{1,2,\dots, T\}$ for discrete-time cases and $t\in[0,T]$ for continuous-time cases. $\bar{a}_t$ and $\bar{b}_t$ are time-dependent coefficients specified by a given noise schedule. The general diffusion models learn the backward process, which is the reverse-time counterpart to the forward diffusion process.

For discrete-time cases, the forward process can be alternatively written as per-step diffusion process as follows. Given $x_0\sim q(x_0)$, we have the forward diffusion process:
\begin{align*}
x_t=a_t x_{t-1} + b_t \varepsilon_{t-1}, \varepsilon_{t-1}\sim\mathcal{N}(0, \mathbf{I})
\end{align*}
with $t\in\{1,2,\dots, T\}$ and existence of the representation of $\bar{a}_t$ and $\bar{b}_t$ by $a_t$ and $b_t$.
\label{def:dm}
\end{definition}

The diffusion process can be characterized as the solution to a stochastic differential equation (SDE), which is commonly expressed in the following general form:
\begin{align}
dx = f(x, t)  dt + g(x, t)  dw,
\label{eq:general_sde}
\end{align}
where $f(x, t)$ is the drift coefficient, $g(x, t)$ is the diffusion coefficient, $w$ is the standard Wiener process.



An instantiation of diffusion models can select different schedules $a_t$ and $b_t$. For example, for a discrete-time diffusion process with a variance preserving schedule $a_t=\sqrt{1-\beta_t}, b_t=\sqrt{\beta_t}$, the corresponding SDE is:
\begin{align*}
 dx= -\frac{1}{2}\beta_t x dt + \sqrt{\beta_t} dw
\end{align*}

In diffusion models, there's also often an interest in the backward SDE for sampling, which has the following form for general SDE as Eq.~\eqref{eq:general_sde}~\cite{anderson1982reverse}:
\begin{align}
dx = \big[ f(x, t) - \nabla_x \cdot [g(x,t)g(x,t)^\intercal] - g(x,t)g(x,t)^\intercal \nabla_{x} \log p_t(x) \big] dt + g(x, t) \, d\bar{w},
\label{eq:pf_sde}
\end{align}
where $p_t(x)$ is the probability distribution of $x_t$ at diffusion time $t$, and $\bar{w}$ is a reverse-time Wiener process.

For above SDE, there exists an ordinary differential equation (ODE), with equivalent marginal distribution $p_t(x)$~\cite{song2021score}:
\begin{align*}
dx = \big[f(x,t) - \frac{1}{2}\nabla_x \cdot [g(x,t)g(x,t)^\intercal] - \frac{1}{2} g(x,t)g(x,t)^\intercal \nabla_x \log p_t(x) \big]dt
\end{align*}

In Eq.~\eqref{eq:pf_sde}, when $g(x,t)$ does not depend on $x$, as $g(t)$ in diffusion models, we have a simplified form:
\begin{align}
dx = \big[ f(x, t) - g(t)g(t)^\intercal \nabla_{x} \log p_t(x) \big] dt + g(t) \, d\bar{w},
\label{eq:dm_reverse_sde}
\end{align}
since the divergence $\nabla_x \cdot [g(\cdot, t)g(\cdot, t)^\intercal]=0$. The corresponding simplified reverse ODE is:
\begin{align}
dx = \big[f(x,t) - \frac{1}{2} g(t)g(t)^\intercal \nabla_x \log p_t(x) \big]dt
\label{eq:dm_reverse_ode}
\end{align}
It is observed that estimating the score function, $\nabla \log p_t(x)$, is a crucial requirement in the reverse sampling process, serving as one of the fundamental components of diffusion models.

\subsection{Diffusion Model}
Diffusion models are a class of generative models that iteratively transform simple noise distributions into complex data distributions through a sequence of forward and reverse processes. This section offers a detailed introduction to the derivation of diffusion model fundamentals, training objectives, and inference mechanisms.

\subsubsection{Foundamentals}
Suppose the data is sampled from distribution $x_0 \sim q(x_0)$, the diffusion model is trained to generate samples $\hat{x}_0$ that follow the same distribution, starting from random Gaussian noise. Three important variants are introduced in this section, including denoising diffusion probabilistic model, denoising diffusion implicit model and score-based denoising diffusion.

\paragraph{Denoising Diffusion Probabilistic Model~\cite{ho2020denoising}.}

The distribution $q(x_t|x_{t-1})=\mathcal{N}(x_t; \sqrt{\alpha_t}x_{t-1}, (1-\alpha_t)\mathbf{I})$ can be written as:
\begin{align}
\Aboxed{x_t=\sqrt{\alpha_t}x_{t-1}+\sqrt{1-\alpha_t}\varepsilon, \varepsilon\sim\mathcal{N}(0, \mathbf{I})}
    \label{eq:x_t}
\end{align}
$\beta_t=1-\alpha_t$ is the variance of noise added at each step $t\in[T]$.

By chain rule, 
\begin{align}x_t=\sqrt{\bar{\alpha}_t}x_0+\sqrt{1-\bar{\alpha}_t}\varepsilon_0, \varepsilon_0\sim\mathcal{N}(0, \mathbf{I})
    \label{eq:xt_x0}
\end{align}
with $\bar{\alpha}_t=\Pi_{i=1}^t \alpha_i$. Equivalently, we have $x_t\sim q(x_t|x_0)=\mathcal{N}(x_t;\sqrt{\bar{\alpha}_t}x_0, (1-\bar{\alpha}_t)\mathbf{I})$.
This equation is also used to predict:
\begin{align}
\Aboxed{\hat{x}_0=\frac{1}{\sqrt{\bar{\alpha}_t}}x_t-\frac{\sqrt{1-\bar{\alpha}_t}}{\sqrt{\bar{\alpha}_t}}\epsilon_\theta}
\label{eq:tweedie}
\end{align}

which is called the Tweedie's formula. $\epsilon_\theta$ is the approximated prediction of $\varepsilon_0$ with a parameterized model by $\theta$, conditioning on $x_t$ and $t$:
\begin{equation*}
\epsilon_\theta(x_t, t) \approx \mathbb{E}_{X_0\sim q}\left[\left.\frac{1}{\sqrt{1-\bar{\alpha}_t}}X_t - \frac{\sqrt{\bar{\alpha}_t}}{\sqrt{1-\bar{\alpha}_t}}X_0 \right|X_t = x_t\right]
\end{equation*}
where $X_0, X_1$ are generic random variables.
The function $\epsilon_\theta(x_t, t)$ consistently takes $x_t$ and $t$ as conditions; however, for simplicity, these dependencies are omitted in subsequent sections of the paper, like in Eq.~\eqref{eq:tweedie}. 
We intentionally use distinct notations, $\epsilon_\theta$ and $\varepsilon$, throughout the paper. Here, $\varepsilon \sim \mathcal{N}(0, \mathbf{I})$ represents true random variables, while $\epsilon_\theta$ denotes their parameterized predictions. Despite both representing noise in the diffusion process, this distinction highlights their respective roles.

The forward diffusion process, as described above, corresponds to the forward stochastic differential equation (SDE) in Eq.\eqref{eq:general_sde}. Our focus also extends to the reverse diffusion process, commonly referred to as the denoising process, which aligns with the reverse SDE in Eq.\eqref{eq:pf_sde}. In discrete-time scenarios, such as those in DDPM, the posterior probability $q(x_{t-1}|x_t, x_0)$ for a one-step denoising process has an explicit analytical form, which is derived below.

By Bayes' rule, we have the posterior distribution,
\begin{align}
    q(x_{t-1}|x_t, x_0)&=\frac{q(x_{t-1}, x_t| x_0)}{q(x_t|x_0)} = \frac{q(x_t|x_{t-1}, x_0)q(x_{t-1}|x_0)}{q(x_t|x_0)}\nonumber \\
    &\propto \mathcal{N}\big(x_{t-1}; \frac{\sqrt{\alpha_t}(1-\bar{\alpha}_{t-1})x_t+\sqrt{\bar{\alpha}_{t-1}}(1-\alpha_t)x_0}{1-\bar{\alpha}_t}, \frac{(1-\alpha_t)(1-\bar{\alpha}_{t-1})}{1-\bar{\alpha}_t}\mathbf{I}\big)
    \label{eq:q_x_t_1}
\end{align}
where we assume $q(x_{t-1} | x_t, x_0) = q(x_{t-1} | x_t) $ as a Markov chain with independent random noise $\varepsilon$ at each timestep. In above Gaussian distribution, 
$\mu(x_t, x_0)=\frac{\sqrt{\alpha_t}(1-\bar{\alpha}_{t-1})x_t+\sqrt{\bar{\alpha}_{t-1}}(1-\alpha_t)x_0}{1-\bar{\alpha}_t}$, plug $x_0=\frac{1}{\sqrt{\bar{\alpha}_t}}x_t-\frac{\sqrt{1-\bar{\alpha}_t}}{\sqrt{\bar{\alpha}_t}}\varepsilon_0$ (derived from Eq.~\eqref{eq:xt_x0}) into it, we have the posterior mean and variance:
\begin{align*}
    \Aboxed{\mu(x_t, \varepsilon_0)&=\frac{1}{\sqrt{\alpha_t}}x_t-\frac{1-\alpha_t}{\sqrt{1-\bar{\alpha}_t}\sqrt{\alpha_t}}\varepsilon_0}\\
    \Aboxed{\sigma_t^2&=\frac{(1-\alpha_t)(1-\bar{\alpha}_{t-1})}{1-\bar{\alpha}_t}}
\end{align*}
which is also used for reparameterization between $\epsilon_\theta$ and $\mu_\theta$. $\epsilon_\theta$ and $\mu_\theta$ are approximated prediction of $\varepsilon_0$ and $\mu(x_t, \varepsilon_0)$ with parameterized models by $\theta$.

Therefore, by replacing $\varepsilon_0$ with $\epsilon_\theta(x_t, t)$, $q(x_{t-1}|x_t, x_0)$ can also be written as:
\begin{align}
    \Aboxed{x_{t-1}(x_t, \epsilon_\theta, \varepsilon)=\frac{1}{\sqrt{\alpha_t}}x_t-\frac{1-\alpha_t}{\sqrt{1-\bar{\alpha}_t}\sqrt{\alpha_t}}\epsilon_\theta+\sigma_t \varepsilon}
    \label{eq:x_t_1}
\end{align}
which is the Langevin dynamics sampling from $x_t$ to $x_{t-1}$, with $\sigma_t^2=\frac{(1-\alpha_t)(1-\bar{\alpha}_{t-1})}{1-\bar{\alpha}_t}, \varepsilon\sim\mathcal{N}(0, \mathbf{I})$. $\sigma_t^2$ is sometimes also set to just $1-\alpha_t$~\cite{ho2020denoising} due to similar experimental performances.

Plug Eq.~\eqref{eq:tweedie} into Eq.~\eqref{eq:x_t_1},
\begin{align}
    x_{t-1}(x_t, \hat{x}_0, \varepsilon)&=\frac{1}{\sqrt{\alpha_t}}x_t-\frac{1-\alpha_t}{\sqrt{1-\bar{\alpha}_t}\sqrt{\alpha_t}}[\frac{x_t-\sqrt{\bar{\alpha}_t}\hat{x}_0}{\sqrt{1-\bar{\alpha}_t}}]+\sigma_t \varepsilon \nonumber\\
    \Leftrightarrow \Aboxed{x_{t-1}(x_t, \hat{x}_0, \varepsilon)&=\frac{\sqrt{\bar{\alpha}_{t-1}}(1-\alpha_t)}{1-\bar{\alpha}_t}\hat{x}_0+\frac{\sqrt{\alpha_t}(1-\bar{\alpha}_{t-1})}{1-\bar{\alpha}_t}x_t+\sigma_t \varepsilon}
    \label{eq:x_t_1_2}
\end{align}

Note that Eq.~\eqref{eq:x_t_1} is also equivalently written as the following in DDIM paper~\cite{song2020denoising}:
\begin{align}
    \Aboxed{x_{t-1}(\hat{x}_0, \epsilon_\theta, \varepsilon)&=\sqrt{\bar{\alpha}_{t-1}}\hat{x}_0 + \sqrt{1-\bar{\alpha}_{t-1}-\sigma_t^2}\epsilon_\theta + \sigma_t \varepsilon}
    \label{eq:x_t_1_ddim}\\
    \Leftrightarrow \Aboxed{x_{t-1}(x_t, \epsilon_\theta, \varepsilon)&=\sqrt{\bar{\alpha}_{t-1}}\big(\frac{x_t-\sqrt{1-\bar{\alpha}_t}\epsilon_\theta}{\sqrt{\bar{\alpha}_t}}\big) + \sqrt{1-\bar{\alpha}_{t-1}-\sigma_t^2}\epsilon_\theta + \sigma_t \varepsilon}
    \label{eq:x_t_1_3}
\end{align}
with $\hat{x}_0=\frac{x_t-\sqrt{1-\bar{\alpha}_t}\epsilon_\theta(x_t, t)}{\sqrt{\bar{\alpha}_t}}$ as the prediction of $x_0$ given $(x_t, t)$. The equivalence of Eq.~\eqref{eq:x_t_1} and above equations can be proved easily as follows.
We have,
\begin{align*}
    \sqrt{1-\bar{\alpha}_{t-1}-\sigma_t^2}&=\sqrt{1-\bar{\alpha}_{t-1}-\frac{1-\bar{\alpha}_{t-1}}{1-\bar{\alpha}_t}(1-\alpha_t)}\\
    &=(1-\bar{\alpha}_{t-1})\sqrt{\frac{\alpha_t}{1-\bar{\alpha}_t}}
\end{align*}
Therefore, with Eq.~\eqref{eq:x_t_1_ddim},
\begin{align*}
    x_{t-1}(\hat{x}_0, \epsilon_\theta, \varepsilon)&=\sqrt{\bar{\alpha}_{t-1}}\hat{x}_0 + \sqrt{1-\bar{\alpha}_{t-1}-\sigma_t^2}\epsilon_\theta + \sigma_t \varepsilon\\
    &=\sqrt{\bar{\alpha}_{t-1}}\hat{x}_0 + \sqrt{1-\bar{\alpha}_{t-1}-\sigma_t^2}\epsilon_\theta+ \sigma_t \varepsilon\\
    &=\sqrt{\bar{\alpha}_{t-1}}\hat{x}_0 + (1-\bar{\alpha}_{t-1})\sqrt{\frac{\alpha_t}{1-\bar{\alpha}_t}}\frac{x_t-\sqrt{\bar{\alpha}_t}\hat{x}_0}{\sqrt{1-\bar{\alpha_t}}}+ \sigma_t \varepsilon\\
   \Leftrightarrow x_{t-1}(\hat{x}_0, x_t, \varepsilon)&=\frac{\sqrt{\bar{\alpha}_{t-1}}(1-\alpha_t)}{1-\bar{\alpha}_t}\hat{x}_0+\frac{\sqrt{\alpha_t}(1-\bar{\alpha}_{t-1})}{1-\bar{\alpha}_t}x_t+\sigma_t \varepsilon
\end{align*}
Therefore, equivalence of Eq.~\eqref{eq:x_t_1_ddim} and Eq.~\eqref{eq:x_t_1_2} is proved.

From Eq.~\eqref{eq:x_t_1_ddim}, another form of $x_{t-1}$ is as following:
\begin{align}
    x_{t-1}(\hat{x}_0, \epsilon_\theta, \varepsilon)&=\sqrt{\bar{\alpha}_{t-1}}\hat{x}_0 + \sqrt{1-\bar{\alpha}_{t-1}-\sigma_t^2}\epsilon_\theta+\sigma_t \varepsilon \nonumber\\
    &=\sqrt{\bar{\alpha}_{t-1}}\hat{x}_0 + \sqrt{1-\bar{\alpha}_{t-1}-\sigma_t^2}(\frac{1}{\sqrt{1-\bar{\alpha}_t}}x_t - \frac{\sqrt{\bar{\alpha}_t}}{\sqrt{1-\bar{\alpha}_t}}\hat{x}_0)+\sigma_t \varepsilon \nonumber\\
   \Leftrightarrow \Aboxed{x_{t-1}(x_t, \hat{x}_0, \varepsilon)&=(\sqrt{\bar{\alpha}_{t-1}}-\sqrt{1-\bar{\alpha}_{t-1}-\sigma_t^2}\frac{\sqrt{\bar{\alpha}_t}}{\sqrt{1-\bar{\alpha}_t}})\hat{x}_0 + \frac{\sqrt{1-\bar{\alpha}_{t-1}}}{\sqrt{1-\bar{\alpha}_t}}x_t+\sigma_t \varepsilon}
    \label{eq:x_t_1_4}
\end{align}

Therefore, Eq.~\eqref{eq:x_t_1_2}, Eq.~\eqref{eq:x_t_1_ddim}, Eq.~\eqref{eq:x_t_1_3} and Eq.~\eqref{eq:x_t_1_4} are all equivalent forms of Eq.~\eqref{eq:x_t_1} for deriving $x_{t-1}$ with $x_t$, just with different input arguments and formats. It is common to see each representation in practical implementation.

\paragraph{Denoising Diffusion Implicit Model~\cite{song2020denoising}.}
DDIM follows Eq.~\eqref{eq:x_t_1_3} for posterior sampling, with two additional modifications:

\textbf{(i).} DDIM also generalizes for arbitrary subsequence of $\{0, 1,\dots, T\}$ as:
\begin{align}
    \Aboxed{x_{t_{n-1}}(x_{t_n}, \epsilon_\theta, \varepsilon)=\sqrt{\bar{\alpha}_{t_{n-1}}}\big(\frac{x_{t_n}-\sqrt{1-\bar{\alpha}_{t_n}}\epsilon_\theta}{\sqrt{\bar{\alpha}_{t_n}}}\big) + \sqrt{1-\bar{\alpha}_{t_{n-1}}-\sigma_{{t_n}}^2}\epsilon_\theta + \sigma_{t_n} \varepsilon}
    \label{eq:x_t_1_3_general}
\end{align}
with the timestep subsequence being $\{t_0, t_1, \dots, t_N\}\subseteq \{0, 1,\dots, T\}, t_0=0$. Note that $\bar{\alpha}_{t_n}=\Pi^{t_n}_{i=0}\alpha_i\neq \Pi^n_{i=0} \alpha_{t_i}$ 

Why cannot we directly reverse Eq.~\eqref{eq:x_t} to get $x_{t-1}=\frac{x_t-\sqrt{1-\alpha_t}\epsilon_\theta}{\sqrt{\alpha_t}}$?
The reason is that it loses the noise $\varepsilon$ added in the iterative sampling process and becomes deterministic sampling for entire denoising path, leading to only one sample in the end.

\textbf{(ii).} Actually DDIM generalizes $\sigma_t$ in Eq.~\eqref{eq:x_t_1} to be (with additional $\eta$ multiplier):
\begin{align*}
\sigma_t(\eta)=\eta\sqrt{\frac{(1-\alpha_t)(1-\bar{\alpha}_{t-1})}{1-\bar{\alpha}_t}}
\end{align*}
when $\eta=0$ Eq.~\eqref{eq:x_t_1} becomes $x_{t-1}=\frac{1}{\sqrt{\alpha_t}}x_t-\frac{1-\alpha_t}{\sqrt{1-\bar{\alpha}_t}\sqrt{\alpha_t}}\epsilon_\theta$. This makes DDIM to have deterministic sampling process. When $\eta=1$ it becomes original stochastic DDPM. $\eta\in[0,1]$ tunes the extent from deterministic to stochastic sampling.

It can be noticed that Eq.~\eqref{eq:x_t_1} with $\eta=0$ is very close to reverse Eq.~\eqref{eq:x_t}: $x_{t-1}=\frac{1}{\sqrt{\alpha_t}}x_t-\frac{\sqrt{1-\alpha_t}}{\sqrt{\alpha_t}}\epsilon_\theta$. These two formulas are different because the previous formula is calculating $\mathbb{E}[x_{t-1}|x_t, x_0]$ as the conditional mean of $x_{t-1}$ given $x_t, x_0$, and the latter formula is computing $x_{t-1}$ given $x_t$ assuming a single step noise is $\epsilon_\theta$. 

Another denoising inference example is that consistency model (introduced later in Sec.~\ref{sec:consistency_model}) applies Eq.~\eqref{eq:tweedie} to directly get deterministicly predicted $\hat{x}_0$, but it is further made stochastic by adding noise again using forward diffusion process $x_{t_{n-1}}=\sqrt{\bar{\alpha}_{t_{n-1}}}\hat{x}_0+\sqrt{1-\bar{\alpha}_{t_{n-1}}}\varepsilon$, and so on so forth. The detailed inference process of the consistency model is discussed in Sec.~\ref{sec:consist_infer}. For the usage of subscript in timestep like $t_n$, it is because consistency model follows a continuous time schedule $\{t_n\}_{n=1}^N$, which is different from the discrete time schedule of $t\in\{0, \dots, T\}$ in the diffusion model.

\paragraph{Score-based Denoising Diffusion~\cite{song2020score}.}

Different from above probabilistic inference on a Markov chain, score-based denoising diffusion models approach the diffusion generative process from a score-matching perspective. As outlined in the reverse SDE Eq.~\eqref{eq:dm_reverse_sde} or reverse ODE as Eq.~\eqref{eq:dm_reverse_ode}, the score function $\nabla_{x_t}\log q(x_t)$ represents the gradient of the probability density in the diffusion process. This score function is utilized in the backward denoising process to generate samples from the target data distribution.

The score function for $q(x_t|x_0)$ in Eq.~\eqref{eq:xt_x0} is as following:
\begin{align}
    s(x_t, t)&\coloneq \nabla_{x_t}\log q(x_t) \nonumber\\
    &=\mathbb{E}_{q(x_0)}[\nabla_{x_t}\log q(x_t|x_0)]\nonumber\\
    &=\mathbb{E}_{q(x_0)}[-\frac{\varepsilon}{\sqrt{1-\bar{\alpha}_t}}]\label{eq:score_with_noise}\\
    &=-\frac{\varepsilon}{\sqrt{1-\bar{\alpha}_t}}\\
    &=-\frac{x_t-\sqrt{\bar{\alpha}_t}x_0}{1-\bar{\alpha}_t}
    \quad\quad  \text{(By Eq.~\eqref{eq:xt_x0})}\label{eq:score_x_t} 
\end{align}
The Eq.~\eqref{eq:score_with_noise} applies for a fixed value of random noise $\varepsilon$. The derivation leverages the fact that:
For $x\sim p(x)=\mathcal{N}(\mu, \sigma^2 \mathbf{I})$, $\nabla_x \log p(x)=\nabla_x(-\frac{1}{2\sigma^2}(x-\mu)^2)=-\frac{x-\mu}{\sigma^2}=-\frac{\varepsilon}{\sigma}, \varepsilon\sim\mathcal{N}(0, \mathbf{I})$.
The score function can be approximated with parameterized model as $s_\theta(x_t,t)\approx s(x_t,t)$ in practice, with expression:
\begin{align}
\Aboxed{s_\theta(x_t, t)&=-\frac{\epsilon_\theta(x_t, t)}{\sqrt{1-\bar{\alpha}_t}}}  \label{eq:score_epsilon}
\end{align}
$\epsilon_\theta(x_t, t)$ is the parameterized noise prediction function with time and sample $x_t$ as inputs.

This score function connects DDPM and score-based denoising diffusion, which will be discussed more in later sections. 

\begin{tcolorbox}[colback=gray!10, colframe=gray!50, boxrule=0.5mm, width=\textwidth]
\textbf{Demystify notations from different papers.}

DDPM uses notations as Eq.~\eqref{eq:x_t}: $q(x_t|x_{t-1})=\mathcal{N}(x_t; \sqrt{\alpha^\text{DDPM}_t}x_{t-1}, (1-\alpha^\text{DDPM}_t)\mathbf{I})$. 

DDIM~\cite{song2020denoising} follows the notations of score-based denoising diffusion~\cite{song2020score}, where $q(x_t|x_0)=\mathcal{N}(x_t; \sqrt{\alpha^\text{DDIM}_t}x_0, (1-\alpha^\text{DDIM}_t)\mathbf{I})$. 

LCM~\cite{luo2023latent} and DMD~\cite{yin2024one} use notations $q(x_t|x_0)=\mathcal{N}(x_t; \alpha^\text{LCM}_t x_0, (\sigma^\text{LCM}_t)^2)\mathbf{I})$. 

The connection of three sets of notations is $\bar{\alpha}^\text{DDPM}=\alpha^\text{DDIM}=(\alpha^\text{LCM})^2, \sigma^\text{LCM}=\sqrt{1-(\alpha^\text{LCM})^2}=\sqrt{1-\bar{\alpha}^\text{DDPM}}$. 

This paper keeps DDPM notations, which are more consistent with code implementation.
\end{tcolorbox}


\subsubsection{Training}
The training process of diffusion models centers on deriving an appropriate objective $\mathcal{L}(\theta)$, which can be derived from a variety of foundational perspectives. In most practical cases, $\theta$ represents the parameters of a neural network, optimized using standard machine learning techniques, such as supervised learning with gradient descent. This subsection examines the mathematical formulation of diffusion model training objectives, emphasizing their connections to noise prediction and score matching, while aiming to present a unified perspective on these methodologies.

\label{sec:diff_train}
\paragraph{DDPM Objective for Training.}

We first construct the variational lower bound for $\log p_\theta(x_0)$:\footnote{This part is adopted from Lilian Weng's blog~\cite{weng2021diffusion}, with full credit attributed to it.}
\begin{align*}
- \log p_{\theta}({x}_0) &\leq - \log p_{\theta}({x}_0) + D_{\text{KL}} \left( q({x}_{1:T}|{x}_0) \| p_{\theta}({x}_{1:T}|{x}_0) \right) \\
&= - \log p_{\theta}({x}_0) + \mathbb{E}_{{x}_{1:T} \sim q({x}_{1:T}|{x}_0)} \left[ \log \frac{q({x}_{1:T}|{x}_0)}{p_{\theta}({x}_{0:T})/p_{\theta}({x}_0)} \right] \\
&= - \log p_{\theta}({x}_0) + \mathbb{E}_{q} \left[ \log \frac{q({x}_{1:T}|{x}_0)}{p_{\theta}({x}_{0:T})} + \log p_{\theta}({x}_0) \right] \\
&= \mathbb{E}_{q} \left[ \log \frac{q({x}_{1:T}|{x}_0)}{p_{\theta}({x}_{0:T})} \right]
\end{align*}
which is the variational lower bound of log-likelihood, following a similar derivation as variational auto-encoder~\cite{kingma2013auto} but on a Markov chain.

Next step is to expand the lower bound for variance reduction, with details in \cite{sohl2015deep}:
\begin{align*}
\mathcal{L}_{\text{VLB}} &= \mathbb{E}_{q({x}_{0:T})} \left[ \log \frac{q({x}_{1:T}|{x}_0)}{p_{\theta}({x}_{0:T})} \right] \\
&= \mathbb{E}_{q} \left[ \log \frac{\prod_{t=1}^{T} q({x}_t | {x}_{t-1})}{p_{\theta}({x}_T) \prod_{t=1}^{T} p_{\theta}({x}_{t-1} | {x}_t)} \right] \\
&= \mathbb{E}_{q} \left[ - \log p_{\theta}({x}_T) + \sum_{t=1}^{T} \log \frac{q({x}_t | {x}_{t-1})}{p_{\theta}({x}_{t-1} | {x}_t)} \right] \\
&= \mathbb{E}_{q} \left[ - \log p_{\theta}({x}_T) + \sum_{t=2}^{T} \log \frac{q({x}_t | {x}_{t-1}, {x}_0)}{p_{\theta}({x}_{t-1} | {x}_t)} + \log \frac{q({x}_1 | {x}_0)}{p_{\theta}({x}_0 | {x}_1)} \right] \\
&= \mathbb{E}_{q} \left[ - \log p_{\theta}({x}_T) + \sum_{t=2}^{T} \log \big( \frac{q({x}_{t-1} | {x}_t, {x}_0)}{p_{\theta}({x}_{t-1} | {x}_t)} \cdot \frac{q({x}_t | {x}_0)}{q({x}_{t-1} | {x}_0)}\big) + \log \frac{q({x}_1 | {x}_0)}{p_{\theta}({x}_0 | {x}_1)} \right] \\
&= \mathbb{E}_{q} \left[ - \log p_{\theta}({x}_T) + \sum_{t=2}^{T} \log \frac{q({x}_{t-1} | {x}_t, {x}_0)}{p_{\theta}({x}_{t-1} | {x}_t)} + \sum_{t=2}^T\log \frac{q({x}_t | {x}_0)}{q({x}_{t-1}|{x}_0)} + \log \frac{q({x}_1 | {x}_0)}{p_{\theta}({x}_0 | {x}_1)} \right] \\
&= \mathbb{E}_{q} \left[ - \log p_{\theta}({x}_T) + \sum_{t=2}^{T} \log \frac{q({x}_{t-1} | {x}_t, {x}_0)}{p_{\theta}({x}_{t-1} | {x}_t)} + \log \frac{q({x}_T | {x}_0)}{q({x}_1|{x}_0)} + \log \frac{q({x}_1 | {x}_0)}{p_{\theta}({x}_0 | {x}_1)} \right] \\
&= \mathbb{E}_{q} \left[ \log \frac{q({x}_T|{x}_0)}{p_\theta({x}_T)} + \sum_{t=2}^{T} \log \frac{q({x}_{t-1} | {x}_t, {x}_0)}{p_{\theta}({x}_{t-1} | {x}_t)} - \log p_\theta({x}_0|{x}_1) \right] \\
&= \mathbb{E}_{q} \left[ \underbrace{D_{\text{KL}} \left( q({x}_T | {x}_0) \| p_{\theta}({x}_T) \right)}_{\mathcal{L}_T} + \sum_{t=2}^{T} \underbrace{D_{\text{KL}} \left( q({x}_{t-1} | {x}_t, {x}_0) \| p_{\theta}({x}_{t-1} | {x}_t) \right)}_{\mathcal{L}_{t-1}} \underbrace{- \log p_{\theta}({x}_0 | {x}_1)}_{\mathcal{L}_0} \right]
\end{align*}
Since ${x}_T\sim\mathcal{N}(0, \mathbf{I})$ which is a Gaussian noise, the $\mathcal{L}_T$ term actually does not contain $\theta$ and can be dropped. $\mathcal{L}_0$ is practically usually also dropped or approximated with $D_\text{KL}(q(x_\epsilon|x_1, x_0)||p_\theta(x_\epsilon|x_1))$ with a sufficiently small $\epsilon$ close to zero. Then the entire loss $\mathcal{L}_\text{VLB}$ becomes a sum of KL-divergence between Gaussians, which can be analytically calculated as:
\begin{align*}
\mathcal{L}_t &= \mathbb{E}_{{x}_0, {\varepsilon}_t} \left[ \frac{1}{2|\!|\mathbf{\Sigma}_\theta({x}_t, t)|\!|_2^2} \left|\!\left| {\mu}_t({x}_t, {x}_0) - \mu_\theta({x}_t, t) \right|\!\right|_2^2 \right] \\
&= \mathbb{E}_{{x}_0, {\varepsilon}_t} \left[ \frac{1}{2|\!|\mathbf{\Sigma}_\theta|\!|_2^2} \left|\!\left| \frac{1}{\sqrt{\alpha_t}} \left( {x}_t - \frac{1 - \alpha_t}{\sqrt{1 - \bar{\alpha}_t}} {\varepsilon}_t \right) - \frac{1}{\sqrt{\alpha_t}} \left( {x}_t - \frac{1 - \alpha_t}{\sqrt{1 - \bar{\alpha}_t}} {\epsilon}_\theta({x}_t, t) \right) \right|\!\right|_2^2 \right] \\
&= \mathbb{E}_{{x}_0, {\varepsilon}_t} \left[ \frac{(1 - \alpha_t)^2}{2\alpha_t (1 - \bar{\alpha}_t) |\!|\mathbf{\Sigma}_\theta|\!|_2^2} \left|\!\left| {\varepsilon}_t - {\epsilon}_\theta({x}_t, t) \right|\!\right|_2^2 \right] \\
&= \mathbb{E}_{{x}_0, {\varepsilon}_t} \left[ \frac{(1 - \alpha_t)^2}{2\alpha_t (1 - \bar{\alpha}_t) |\!|\mathbf{\Sigma}_\theta|\!|_2^2} \left|\!\left| {\varepsilon}_t - {\epsilon}_\theta \left( \sqrt{\bar{\alpha}_t} {x}_0 + \sqrt{1 - \bar{\alpha}_t} {\varepsilon}_t, t \right) \right|\!\right|_2^2 \right]
\end{align*}
by choosing parameterization,
\begin{align}
    \mu_\theta(x_t, t)&=\frac{1}{\sqrt{\alpha_t}}x_t-\frac{1-\alpha_t}{\sqrt{1-\bar{\alpha}_t}\sqrt{\alpha_t}}\epsilon_\theta\nonumber
\end{align}

\paragraph{Score Matching Objective for Training.}
We first answer the question:

\begin{center}
\textit{Why using score matching?}
\end{center}

Score matching originates from 2005 by \cite{hyvarinen2005estimation}. Given the ground-truth sample distribution $q(x)$, the objective is to learn the parameterized probability density model $p_\theta$ as:
\begin{align}
    p(x;\theta)=\frac{1}{Z(\theta)}\exp(-E(x;\theta))\nonumber
\end{align}
where the partition function $Z(\theta)=\int_{x\in\mathbb{R}^d}\exp(-E(x;\theta))$ is generally intractable. Score matching is to bypass the estimation of $Z(\theta)$ in above probability density estimation process.

Specifically, the score function is the gradient of the log-density with respect to sample vector as:
\begin{align*}
    \psi(x; \theta) = \begin{pmatrix}
\frac{\partial \log p(x; \theta)}{\partial x_1} \\
\vdots \\
\frac{\partial \log p(x; \theta)}{\partial x_d}
\end{pmatrix}
= \begin{pmatrix}
\psi_1(x; \theta) \\
\vdots \\
\psi_d(x; \theta)
\end{pmatrix}
= \nabla_x \log p(x; \theta)
\end{align*}
which is not dependent on $Z(\theta)$, and this is the critical property for us to leverage the score matching method.

For score matching, two equivalent objectives have been derived: Explicit Score Matching (ESM) and Implicit Score Matching (ISM), as named later by~\cite{vincent2011connection}.

The objectives are:
\begin{align}
    J_{ESM_q}(\theta) = \mathbb{E}_{q(x)} \left[ \frac{1}{2} \left\| \psi(x; \theta) - \frac{\partial \log q(x)}{\partial x} \right\|^2_2 \right]\nonumber
\end{align}
\begin{align*}
    J_{ISM_q}(\theta) &=\mathbb{E}_{q(\mathbf{x})} \left[ \frac{1}{2} \| \psi(x; \theta) \|^2_2 + \nabla_x \cdot \psi(x;\theta) \right] \quad (\nabla_x \cdot () \text{ as divergence})\\
    &= \mathbb{E}_{q(\mathbf{x})} \left[ \frac{1}{2} \| \psi(x; \theta) \|^2_2 + \sum_{i=1}^{d} \frac{\partial \psi_i(x; \theta)}{\partial x_i} \right] \\
    &=\mathbb{E}_{q(\mathbf{x})} \left[ \frac{1}{2} \| \psi(x; \theta) \|^2_2 + \text{Tr}(\nabla_x \psi(x;\theta)) \right]
\end{align*}
where $\nabla_x\cdot \psi(x;\theta) = \text{Tr}(\nabla_x \psi(x;\theta))$ as $\psi:\mathbb{R}^d\rightarrow \mathbb{R}^d$ is a vector field.

We have equivalence of ESM and ISM:
\begin{align}
    J_{ESM_q}(\theta) =J_{ISM_q}(\theta) + C\nonumber
\end{align}
where $C$ is a constant independent on $\theta$.

Score-based denoising diffusion~\cite{song2019generative, song2020score} applies ESM objective generalized for $t\in[T]$, where the score function $\nabla_{x_t}\log q(x_t)$ can be estimated with neural networks $s_\theta$:
\begin{align*}
    \mathcal{L}_\text{SM}&=\int_0^T w(t)\mathbb{E}_{x_t\sim q(x_t)} [||\nabla_{x_t} \log q(x_t) - s_\theta(x_t, t)||_2^2] \mathrm{d}t\\
    &\approx \mathbb{E}_{t\in[0, T],x_t\sim q(x_t)} [w(t)||\nabla_{x_t} \log q(x_t) - s_\theta(x_t, t)||_2^2] \quad \text{(time discretization)} \\
    &=\mathbb{E}_{t\in[0, T],x_t\sim q(x_t)}[w(t)||-\frac{x_t-\sqrt{\bar{\alpha}_t}x_0}{1-\bar{\alpha}_t} + \frac{\epsilon_\theta(x_t, t)}{\sqrt{1-\bar{\alpha}_t}}||_2^2] \quad \text{(by Eq.~\eqref{eq:score_x_t} and \eqref{eq:score_epsilon} )}\\
    &=\mathbb{E}_{x_0\sim q(x_0), t\in[0, T], \varepsilon\sim \mathcal{N}(0, \mathbf{I})}[\frac{w(t)}{\sqrt{1-\bar{\alpha}_t}}||\epsilon_\theta(x_t, t) - \varepsilon||_2^2]\quad  \text{(By Eq.~\eqref{eq:xt_x0})}
\end{align*}
with $x_t=\sqrt{\bar{\alpha}_t}x_0+\sqrt{1-\bar{\alpha}_t}\varepsilon, \varepsilon\sim \mathcal{N}(0, \mathbf{I})$. $\epsilon_\theta$ is the parameterized noise prediction model for $\epsilon$-prediction. Other forms of prediction are discussed in Sec.~\ref{sec:parameterization}. This matches the loss used in DDPM regardless of the time-dependent coefficients. 

The naive choice of $w(t)=1$ and constant $t$ leads to that the score matching norm is estimated on the sample distribution $\mathbb{E}_{x_t\sim q(x_t)}[||\nabla_x \log q(x_t)- s_\theta(x_t)||_2^2]$, which causes inaccurate score estimation for low density regions in the sample space. There is also the trade-off of the noise scale (depending on $t$) added to $x_t$, that smaller noise leads to lower coverage of low density regions and less corruption of data distribution, while larger noise leads to higher coverage of low density regions and more corruption of the data distribution. 

Later improvement applies multi-scale noise perturbation with noise-conditional score network (NCSN)~\cite{song2019generative, song2020improved} $s_\theta(x_t, t)$ to solve this issue, with the loss expression shown in above $\mathcal{L}_\text{SM}$ equation. $w(t)$ is typically chosen to be $w(t)\propto 1/\mathbb{E}[||\nabla_{x_t}\log q(x_t|x_0)||^2]$ for loss balancing over time. For example, it can be set as $w(t)=\sigma^2_t$.

\begin{tcolorbox}[colback=gray!10, colframe=gray!50, boxrule=0.5mm, width=\textwidth]
\textbf{(Approximate) equivalence of objectives.}

Regardless of the coefficients, the score matching objective coincides with the DDPM objective as the mean-squared-error (MSE) of noise prediction. In practice, the coefficients in front of MSE are usually dropped for simplicity.
\end{tcolorbox}

\subsubsection{Inference}
\label{sec:diff_inf}
After training, the diffusion models are applied for the denoising function in an iterative manner to generate diverse yet accurate prediction of $\hat{x}_0$, which is referred to as the inference process.
We describe three types of sampling process for diffusion model inference.

\textbf{(1). DDPM Sampling:}
The inference process of DDPM (for $\epsilon$-prediction model) requires iteratively: (i). predicting noise $\epsilon_\theta(x_t, t)$, starting with Gaussian noise $x_T\sim\mathcal{N}(0, \mathbf{I})$; (ii). derive $\hat{x}_0$ using $\epsilon_\theta$ with $\hat{x}_0=\frac{1}{\sqrt{\bar{\alpha}_t}}x_t-\frac{\sqrt{1-\bar{\alpha}_t}}{\sqrt{\bar{\alpha}_t}}\epsilon_\theta$ by Eq.~\eqref{eq:tweedie};
(iii). 
$x_{t-1}(x_t, \hat{x}_0, \varepsilon)=\frac{\sqrt{\bar{\alpha}_{t-1}}(1-\alpha_t)}{1-\bar{\alpha}_t}\hat{x}_0+\frac{\sqrt{\alpha_t}(1-\bar{\alpha}_{t-1})}{1-\bar{\alpha}_t}x_t+\sigma_t \varepsilon$ by Eq.~\eqref{eq:x_t_1_2}. This iterative process is executed along the entire diffusion timestep sequence $[T]$ for original DDPM sampling. Compared with few-step sampling, the multi-step iterative denoising and diffusion process increases the diversity and fidelity of generated samples. The procedure for DDPM inference is shown in Alg.~\ref{alg:eps_ddpm}. Practically, there exists a simpler sampling method for DDPM inference as Alg.~\ref{alg:ddpm_if_simple}, which no longer requires estimation of posterior mean and variance. The Alg.~\ref{alg:ddpm_if_simple} as a simpler form cannot be used to estimate the probability of samples due do lack of mean and standard deviation for the distribution. For both algorithms, it employs the $\epsilon$-prediction as in most practical usage, where the parameterized models directly approximates the noise $\epsilon_\theta$. Other forms of prediction parameterization can be applied with appropriate variable transformations, as discussed in detail in later Sec.~\ref{sec:parameterization}

\textbf{(2). DDIM Sampling:}
Instead of conducting denoising along the entire diffusion process that can be computationally expensive at inference time, DDIM or stratified sampling allows to sample along a sub-set $\{t_n\}_{n=1}^N\subseteq [T]$ for inference acceleration.

The DDIM sampling follows Eq.~\eqref{eq:x_t_1_3_general}, by replacing the $x_{t-1}$-prediction formula in DDPM inference step (iii) with the following:
\begin{align}
    x_{t_{n-1}}(x_{t_n}, \epsilon_\theta, \varepsilon)=\sqrt{\bar{\alpha}_{t_{n-1}}}\big(\frac{x_{t_n}-\sqrt{1-\bar{\alpha}_{t_n}}\epsilon_\theta}{\sqrt{\bar{\alpha}_{t_n}}}\big) + \sqrt{1-\bar{\alpha}_{t_{n-1}}-\sigma_{{t_n}}^2}\epsilon_\theta + \sigma_{t_n} \varepsilon
    \nonumber
\end{align}
for any subsequence $\{t_0, t_1, \dots, t_N\}\subseteq [T], t_0=0$.

A side-by-side comparison of DDPM and DDIM inference is displayed as Alg.~\ref{alg:eps_ddpm} and Alg.~\ref{alg:eps_ddim}. It may be noticed that DDIM uses Eq.~\eqref{eq:x_t_1_3_general} instead of Eq.~\eqref{eq:x_t_1_2} in DDPM as posterior mean prediction, although the two equations are previously proved to be equivalent. The two equations behave differently when the inference is performed on a subsequence $\{t_n\}_{n=1}^N\subseteq [T]$ in DDIM. For example, when $t-1=t_{n-1}=0, \bar{\alpha}_{t-1}=\bar{\alpha}_{t_{n-1}}=1$, Eq.~\eqref{eq:x_t_1_3_general} gives  $\mu=\hat{x}_0$ while Eq.~\eqref{eq:x_t_1_2} gives $\mu=\frac{(1 - \alpha_t)\sqrt{\bar{\alpha}_t}}{1-\bar{\alpha}_t}\hat{x}_0$, which is wrong. This indicates that Eq.~\eqref{eq:x_t_1_2} cannot be applied on a subsequence of $[T]$ as in DDIM. The procedure for DDIM inference is shown in Alg.~\ref{alg:eps_ddim}, with $\epsilon$-prediction explained above.

\textbf{(3). SMLD Sampling:}
Score matching is used for learning the score function, and Langevin dynamics is used for sampling at inference time. The joint uasge is called score matching Langevin dynamics (SMLD)~\cite{song2019generative}.
After learning the score function, new samples can be generated using Langevin dynamics, which is derived by the iterative update rule:
\[
x_{t+1} = x_t + \frac{\delta}{2} s_\theta(x_t) + \sqrt{\delta} \varepsilon_t
\]
where:
\begin{itemize}
    \item $x_0$ is sampled from Gaussian distribution;
    \item $\delta$ is the step size;
    \item $s_\theta(x_t)$ is the learned score function;
    \item $\varepsilon_t \sim \mathcal{N}(0, \mathbf{I})$ is Gaussian noise.
\end{itemize}
Here it follows the original update rules as presented in \cite{song2019generative}, where the process begins with a random Gaussian $x_0$ and progresses towards $x_T$, representing the target distribution. This approach is equivalent to the formulation in DDPM by a change of variable $x_t \rightarrow x_{T-t}$.
In this framework, the denoising process proceeds with increasing $t$, with the sample distribution being derived as $T \rightarrow \infty$. This process iteratively refines the sample $x_t$ with the learned score and incorporating Gaussian noise to ensure adequate exploration. In the limit of step size approaching zero, the process converges to samples from the true distribution through Langevin dynamics.

\begin{figure}[htbp]
\begin{minipage}{0.55\textwidth}
\begin{algorithm}[H]
\caption{DDPM inference with $\epsilon$-prediction}
\begin{algorithmic}[1]
\State $x_T\sim\mathcal{N}(0, \mathbf{I})$
\For{$t \in [T]$}
    \State predict noise $\epsilon_\theta(x_t, t)$
    \State $\hat{x}_0 \gets \frac{1}{\sqrt{\bar{\alpha}_t}}x_t-\frac{\sqrt{1-\bar{\alpha}_t}}{\sqrt{\bar{\alpha}_t}}\epsilon_\theta$ by Eq.~\eqref{eq:tweedie}
    \State $\mu_\theta(x_{t-1})\gets \frac{(1 - \alpha_t)\sqrt{\bar{\alpha}_t}}{1-\bar{\alpha}_t}\hat{x}_0  + \frac{\sqrt{\alpha}_t(1-\bar{\alpha}_{t-1})}{1-\bar{\alpha}_t} x_t$ as Eq.~\eqref{eq:x_t_1_2}
    \State $\sigma_t \gets \sqrt{(1 - \alpha_t)\frac{1-\bar{\alpha}_{t-1}}{1-\bar{\alpha}_t}}$
    \State $x_{t-1} \gets \mu_\theta + \sigma_t \cdot \varepsilon,  \varepsilon \sim \mathcal{N}(0,\mathbf{I})$
\EndFor
\end{algorithmic}
\label{alg:eps_ddpm}
\end{algorithm}
\end{minipage}
\hfill
\begin{minipage}{0.44\textwidth}
\vspace{-1.4cm}
\begin{algorithm}[H]
\caption{DDPM inference with $\epsilon$-prediction (simplified)}
\begin{algorithmic}[1]
\State $x_T\sim\mathcal{N}(0, \mathbf{I})$
\For{$t \in [T]$}
    \State predict noise $\epsilon_\theta(x_t, t)$
    \State $x_{t-1} \gets \frac{1}{\sqrt{\alpha_t}}(x_t-\frac{1-\alpha_t}{\sqrt{1-\bar{\alpha}_t}}\epsilon_\theta)+\sqrt{1-\bar{\alpha}_{t-1}} \varepsilon,  \varepsilon \sim \mathcal{N}(0,\mathbf{I})$ as Eq.~\eqref{eq:x_t_1}
\EndFor
\end{algorithmic}
\label{alg:ddpm_if_simple}
\end{algorithm}
\end{minipage}
\end{figure}

\begin{algorithm}[H]
\caption{DDIM inference with $\epsilon$-prediction}
\begin{algorithmic}[1]
\State $x_{t_N}\sim\mathcal{N}(0, \mathbf{I}), \{t_n\}_{n=1}^N\subseteq[T]$
\For{$n\in[N]$}
    \State predict noise $\epsilon_\theta(x_{t_t}, t_n)$
    \State $\hat{x}_0 \gets \frac{1}{\sqrt{\bar{\alpha}_{t_n}}}x_{t_n}-\frac{\sqrt{1-\bar{\alpha}_{t_n}}}{\sqrt{\bar{\alpha}_{t_n}}}\epsilon_\theta$ by Eq.~\eqref{eq:tweedie}
    \State $\mu_\theta(x_{t_{n-1}})\gets \sqrt{\bar{\alpha}_{t_{n-1}}}\hat{x}_0 + \sqrt{1-\bar{\alpha}_{t_{n-1}}-\sigma_{{t_n}}^2}\epsilon_\theta$ as Eq.~\eqref{eq:x_t_1_3_general}
    \State $\sigma_{t_n} \gets \eta\sqrt{(1 - \alpha_{t_n})\frac{1-\bar{\alpha}_{t_{n-1}}}{1-\bar{\alpha}_{t_n}}}$
    \State $x_{t_{n-1}} \gets \mu_\theta + \sigma_{t_n} \cdot \varepsilon,  \varepsilon \sim \mathcal{N}(0,\mathbf{I})$
\EndFor
\end{algorithmic}
\label{alg:eps_ddim}
\end{algorithm}

\subsection{Consistency Model}
\label{sec:consistency_model}
The diffusion model~\cite{ho2020denoising, song2020score} solves the multi-modal distribution matching problem with a stochastic differential equation (SDE), while the consistency model~\cite{song2023consistency} solves an equivalent probability flow ordinary differential equation (ODE): 
\begin{align}
    \frac{dx_t}{dt}=-t \nabla \log p_t(x) \nonumber
\end{align}
with $p_t(x)=p_\text{data}(x)\otimes \mathcal{N}({0}, t^2\mathbf{I})$ for time period $t\in[0, T]$, where $p_\text{data}(x)$ is the data distribution. The reverse process along the solution trajectory $\{\hat{{x}}_\tau\}_{\tau\in[\epsilon, T]}$ of this ODE is the data generation process from initial random samples $\hat{{x}}_T\sim\mathcal{N}({0}, T^2\mathbf{I})$, with $\epsilon$ as a small constant close to $0$ for handling numerical problem at the boundary. Specifically, it approximates a parameterized consistency function $f_\theta: (x_t, t)\rightarrow x_\epsilon$, which is defined as a map from the noisy sample $x_t$ at step $t$ back to the original sample $x_\epsilon$, instead of applying a step-by-step denoising function $p_\theta(x_{t-1}|x_t)$ as the reverse diffusion process in diffusion model.

\subsubsection{Training}
\label{sec:consist_train}
The parameterized consistency model $f_\theta$ is enforced to satisfy the boundary condition:
\begin{align}
f_\theta(x_t,t) &= c_{\text{skip}}(t) x_t + c_{\text{out}}(t)x_\theta(x_t, c, t)
\label{eq:consist}
\end{align}
where $c_{\text{skip}}(t)=\frac{\sigma^2}{(t-\epsilon)^2+\sigma^2}$ and $c_{\text{out}}(t)=\frac{\sigma(t-\epsilon)}{\sqrt{t^2+\sigma^2}}$, as a boundary condition satisfying $f_\theta(x_\epsilon, \epsilon)=x_\epsilon\approx x_0$ when $\epsilon \rightarrow 0$. In practice the data standard deviation takes $\sigma=0.5$, small constant $\epsilon=0.002$ (practically usually using $t_{-1}$ for indexing as a trick)\footnote{Append $0.002$ to get time sequence $\{t_1, \dots, t_n, \dots, t_N, 0.002\}$, and retrieve value $0.002$ with $n=-1$.}. 

The training process of a consistency model can be achieved with (i). consistency distillation with a ODE solver or (ii). consistency training without the solver.

\paragraph{Consistency Distillation.}
The consistency distillation loss is defined as the $\lambda(t_n)$-weighted distance of two model predictions:
\begin{align}
    \mathcal{L}_\text{CD}(\theta)&=\mathbb{E}[\lambda(t_n)d(f_\theta(x_{t_{n+1}}, t_{n+1}),f_{\theta^-}(\hat{x}_{t_n}, t_{n})]\nonumber\\
    \hat{x}_{t_n}&=\text{Denoise}(x_{t_{n+1}}, t_{n+1}, t_n)
    \label{eq:xt_ode}
\end{align}
where $\lambda(t_n)$ is a time dependent coefficient usually set as constant in practice, and $d(\cdot, \cdot)$ is a distance metric like $l_2$ loss $d(x,y)=||x-y||_2^2$,  Pseudo-Huber loss $d(x,y)=\sqrt{||x-y||_2^2+c^2}-c$ with $c>0$~\cite{song2023improved}, or LPIPS loss~\cite{zhang2018unreasonable}. $\text{Denoise}(x_{t_{n+1}}, t_{n+1}, t_n)$ is the denoising function (\emph{e.g.}, an ODE solver) to denoise samples from $t_{n+1}$ to $t_n$. In the distillation setting, the denoising function can be implemented with a pre-trained teacher diffusion model.  For example, by using pretrained diffusion noise prediction $\epsilon_\phi$, based on DDIM sampling as Eq.~\eqref{eq:x_t_1_3_general}:
\begin{align}
    \text{Denoise}(x_{t_{n+1}}, t_{n+1}, t_n;\epsilon_\phi):=\sqrt{\bar{\alpha}_{t_{n}}}\big(\frac{x_{t_{n+1}}-\sqrt{1-\bar{\alpha}_{t_{n+1}}}\epsilon_\phi}{\sqrt{\bar{\alpha}_{t_{n+1}}}}\big) + \sqrt{1-\bar{\alpha}_{t_{n}}-\sigma_{{t_{n+1}}}^2}\epsilon_\phi + \sigma_{t_{n+1}} \varepsilon
    \label{eq:ode_ddim}
\end{align}
$\phi$ is used for $\epsilon_\phi$ here to distinguish from learnable $\theta$. $\sigma_{t_{n+1}}=0$ for DDIM sampler.

\textbf{Option1: multi-step denoising function.}
It also allows to conduct $m$-step denoising as:
\begin{align}
    \hat{x}_{t_n}&=\text{Denoise}^m(x_{t_{n+m}}, t_{n+m}, t_n)\nonumber\\
    &=\text{Denoise}(\text{Denoise}(\dots\text{Denoise}(x_{t_{n+m}}, t_{n+m}, t_n)))
    \label{eq:multi_step_denoise}
\end{align}
with more computational cost.

\textbf{Option2: CFG augmentation.}
It is an option to use classifier-free guidance (CFG)~\cite{ho2022classifier} augmented prediction with given guidance weight $w$ in the denoising function:
\begin{align}
    \epsilon_\phi(x_{t_{n+1}}, c, t_{n+1}, t_n,  w) = \epsilon_\phi(x_{t_{n+1}}, c, t_{n+1}, t_n) + w(\epsilon_\phi(x_{t_{n+1}}, c, t_{n+1}, t_n)- \epsilon_\phi(x_{t_{n+1}}, \varnothing, t_{n+1}, t_n))
    \label{eq:cfg}
\end{align}
Merging above Eq.~\eqref{eq:cfg} into Eq.~\eqref{eq:ode_ddim} gives DDIM sample with CFG augmentation.

\textbf{Reparametrization from DM to CM.}
Given diffusion model (DM) $\epsilon_\theta$, using Eq.~\eqref{eq:consist} and \eqref{eq:xt_x0}, consistency model $f_\theta$ can be reparameterized by (proposed by LCM~\cite{luo2023latent}):
\begin{align}
    f_\theta(x_t,t) &= c_{\text{skip}} x_t + c_{\text{out}}x_\theta(x_t, c, t)\nonumber\\
    x_\theta&=\frac{x_t-\sqrt{1-\bar{\alpha}_t}\epsilon_\theta(x_t,c,t)}{\sqrt{\bar{\alpha}_t}}
    \label{eq:parameterize_cm_dm}
\end{align}
where $x_\theta$ is the prediction of clean sample $\hat{x}_0$. Note that it also requires the DM and CM to have aligned timesteps $t$ (at least DM timesteps need to cover CM timesteps).


\begin{tcolorbox}[colback=gray!10, colframe=gray!50, boxrule=0.5mm, width=\textwidth]
\paragraph{Benefits of the reparameterization.}
In the distillation process of CM from pretrained DM, this parameterization allows $\epsilon_\theta$ to be initialized with $\epsilon_\phi$ for acheiving the minimal distillation efforts.
\end{tcolorbox}

\textbf{Discussion.}
Note that in LCM paper, they write an equivalent representation of Eq.~\eqref{eq:xt_ode} and \eqref{eq:ode_ddim}:
\begin{align}
    \Psi(x_{t_{n+1}}, t_{n+1}, t_n, c)&=\sqrt{\frac{\bar{\alpha}_{t_n}}{\bar{\alpha}_{t_{n+1}}}}x_{t_{n+1}} - \sigma_{t_n}\big(\frac{\sigma_{t_{n+1}}\cdot \sqrt{\bar{\alpha}}_{t_n}}{\sqrt{\bar{\alpha}}_{t_{n+1}}\cdot\sigma_{t_n}}-1\big)\epsilon_\phi(x_{t_{n+1}}, c, t_{n+1}) - x_{t_{n+1}} \nonumber\\
    &=\sqrt{\bar{\alpha}_{t_{n}}}\big(\frac{x_{t_{n+1}}-\sqrt{1-\bar{\alpha}_{t_{n+1}}}\epsilon_\phi}{\sqrt{\bar{\alpha}_{t_{n+1}}}}\big) + \sqrt{1-\bar{\alpha}_{t_{n}}}\epsilon_\phi- x_{t_{n+1}}\nonumber
\end{align}
and
\begin{align*}
        \hat{x}_{t_n}&=x_{t_{n+1}} + (1+w)\Psi(x_{t_{n+1}}, t_{n+1}, t_n, c)-w\Psi(x_{t_{n+1}}, t_{n+1}, t_n, \varnothing)\\
        &=\text{Denoise}(x_{t_{n+1}}, t_{n+1}, t_n;\epsilon_\phi)
\end{align*}
which is exactly the same as Eq.~\eqref{eq:ode_ddim} when $\sigma_{t_{n+1}}=0$.

\paragraph{Consistency Training.}
Different from above consistency distillation process, the consistency training (CT) loss is defined according to the self-consistency property without distillation from any teacher model:
\begin{align}
    \mathcal{L}_\text{CT}(\theta)&=\mathbb{E}[\lambda(t_n)d(f_\theta(x_{t_{n+1}}, t_{n+1}),f_{\theta^-}(x_{t_n}, t_{n})]
    \label{eq:cm_sample}
\end{align}
with $d(\cdot, \cdot)$ as a distance metric.

The following theorem presents the convergence property of CT towards CD with respect to the the timestep interval.
\begin{theorem}\cite{song2023consistency} Define a maximum of timestep interval $\Delta t=\max_{n\in[N-1]}|t_{n+1}-t_n|$, under certain Lipschitz and smooth conditions, and assuming ground-truth teacher score model in CD, we have
\begin{align*}
\mathcal{L}^N_\text{CD}=\mathcal{L}^N_\text{CT} + o(\Delta t)\\
\mathcal{L}^N_\text{CT}\geq O(\Delta t) \text{ if } \inf \mathcal{L}^N_\text{CD}>0
\end{align*}
\label{theorem:ct}
\end{theorem}
where $\Delta t$ can be a function of training iteration $k$. The first equation holds when $\Delta t$ is sufficiently small due to the usage of Taylor expansion.
It essentially implies that when iteration $k$ becomes large enough and $\Delta t(k)\leq \epsilon$, if $\mathcal{L}^N_\text{CT}\le \epsilon_\text{CT}$, then $\mathcal{L}^N_\text{CD}\le \epsilon_\text{CT}+\epsilon$, and $\epsilon$ is much smaller than $\epsilon_\text{CT}$. This is the reason why CT takes $t_n(k)$ with a decreasing scale as training iteration $k$ increases, which tries to minimize $\mathcal{L}^N_\text{CT}$ and the difference of $\mathcal{L}^N_\text{CD}$ and $\mathcal{L}^N_\text{CT}$ when $|t_{n+1}(k)-t_n(k)|$ becomes infinitely small in theory. 


\subsubsection{Inference}
\label{sec:consist_infer}
The inference process of consistency model involves iteratively:
\begin{enumerate}
\item  predict $\hat{x}_0=f_\theta(x_{t_{n+1}}, t_{n+1})$ as Eq.~\eqref{eq:consist}; 
\item add noise with forward diffusion as Eq.~\eqref{eq:xt_x0} for timestep $t_n$:
\end{enumerate}
\begin{align}
    x_{t_n}=\hat{x}_0 + \sqrt{t_n^2 - \epsilon^2}\varepsilon, \varepsilon\sim\mathcal{N}(0, \mathbf{I}) 
    \label{eq:cm_x_t}
\end{align}
in standard consistency model, $\epsilon$ is a small constant value close to 0 as a lower bound of time. This is a variance exploding noise schedule. 

The pseudo-code for multi-step sampling with consistency models is shown in Alg.~\ref{alg:cm}.

\begin{figure}[bp]
\begin{minipage}{0.49\textwidth}
\vspace{-1.1cm}
\begin{algorithm}[H]
\caption{CM inference with $f$-prediction}
\begin{algorithmic}[1]
\State $x_{t_N}\sim\mathcal{N}(0, \mathbf{I}), \{t_n\}_{n=1}^N\subseteq[T]$
\For{$n\in [N]$}
    \State predict $\hat{x}_0\gets f_\theta(x_{t_n}, t_n)$
    \State $x_{t_{n-1}}=\hat{x}_0 + \sqrt{t_{n-1}^2 - \epsilon^2}\varepsilon, \varepsilon\sim\mathcal{N}(0, \mathbf{I}) $ as Eq.~\eqref{eq:cm_x_t}
\EndFor
\end{algorithmic}
\label{alg:cm}
\end{algorithm}
\end{minipage}
\hfill
\begin{minipage}{0.49\textwidth}
\begin{algorithm}[H]
\caption{LCM inference with $\epsilon$-prediction}
\begin{algorithmic}[1]
\State $x_{t_N}\sim\mathcal{N}(0, \mathbf{I}), \{t_n\}_{n=1}^N\subseteq[T]$
\For{$n\in [N]$}
    \State predict noise $\epsilon_\theta(x_{t_n}, t_n)$
    \State $\hat{x}_0 \gets \frac{1}{\sqrt{\bar{\alpha}_{t_n}}}x_{t_n}-\frac{\sqrt{1-\bar{\alpha}_{t_n}}}{\sqrt{\bar{\alpha}_{t_n}}}\epsilon_\theta$ by Tweedie Eq.~\eqref{eq:tweedie}
    \State $x_{t_{n-1}} \gets \sqrt{\bar{\alpha}_{t_{n-1}}}\hat{x}_0 + \sqrt{1-\bar{\alpha}_{t_{n-1}}} \varepsilon,  \varepsilon \sim \mathcal{N}(0,\mathbf{I})$
\EndFor
\end{algorithmic}
\label{alg:lcm}
\end{algorithm}
\end{minipage}
\end{figure}

If using LCM reparameterization, it follows the same variance preserving noise schedule as DDPM:
\begin{align}
   x_{t_n}=\sqrt{\bar{\alpha}_{t_n}}\hat{x}_0+\sqrt{1-\bar{\alpha}_{t_n}}\varepsilon, \varepsilon\sim\mathcal{N}(0, \mathbf{I}) 
   \label{eq:xt_1_x0}
\end{align}

Additionally, LCM has boundary satisfaction as $f_\theta(x_0, t_0)=x_0$ for $t_0=\epsilon$, and:
\begin{align}
    \hat{x}_0=x_\theta&=\frac{x_{t_{n+1}}-\sqrt{1-\bar{\alpha}_{t_{n+1}}}\epsilon_\theta(x_{t_{n+1}},c,{t_{n+1}})}{\sqrt{\bar{\alpha}_{t_{n+1}}}} \nonumber
\end{align}
Plug into Eq.~\eqref{eq:xt_1_x0}
\begin{align}
    x_{t_n}=\sqrt{\bar{\alpha}_{t_n}}\frac{x_{t_{n+1}}-\sqrt{1-\bar{\alpha}_{t_{n+1}}}\epsilon_\theta(x_{t_{n+1}},c,t_{n+1})}{\sqrt{\bar{\alpha}_{t_{n+1}}}}+\sqrt{1-\bar{\alpha}_{t_{n+1}}}\varepsilon, \varepsilon\sim\mathcal{N}(0, \mathbf{I}) \nonumber
\end{align}
The pseudo-code for multi-step sampling with LCM is shown in Alg.~\ref{alg:lcm}.
It is worth noting that the LCM sampling is exactly the same as DDIM sampling with random noise $\sigma_t=0$ (as Eq.~\eqref{eq:x_t_1_3_general} with $\varepsilon\rightarrow\epsilon_\theta$). This provides another interpretation of the relationship between consistency inference and DDIM inference as follows.
\begin{tcolorbox}[colback=gray!10, colframe=gray!50, boxrule=0.5mm, width=\textwidth]
\begin{proposition}
    DDIM inference is approximately (ignoring boundary satisfaction in consistency model) the consistency inference with LCM reparameterization by just using predicted noise $\epsilon_\theta(x_{t_{n+1}}, c, t_{n+1})$ instead of random noise $\varepsilon\sim\mathcal{N}(0, \mathbf{I})$ in the diffusion process from $\hat{x}_0$ to $x_{t_n}$.
\end{proposition}
\end{tcolorbox}

\textbf{Variance Preserving and Variance Exploding Noise Schedule.}
Different from previously introduced DDPM with $x_t=\sqrt{\alpha_t}x_{t-1}+\sqrt{1-\alpha_t}\varepsilon, \varepsilon\sim\mathcal{N}(0, \mathbf{I})$, which has preserved variance with a constant norm, the standard consistency model has exploding variance $\sigma_d^2+t_n^2$, as indicated by noise adding formula Eq.~\eqref{eq:cm_sample}. Its variance increases over the timestep, and $\sigma_d^2$ is the variance of training samples. LCM follows DDPM noise schedule as Eq.~\eqref{eq:xt_1_x0}, therefore also preserves the variance norm even with consistency distillation.

\subsection{Rectified Flow}
\label{sec:rectified_flow}
Rectified flow~\cite{liu2022flow} or flow matching~\cite{lipman2022flow} is a special class of diffusion models, following certain noise schedules in general diffusion models defined as Def.~\ref{def:dm}. For the continuous case, it takes $\bar{a}_t=1-t, \bar{b}_t=t$.

\subsubsection{Training}
\label{subsec:rf_train_infer}
\paragraph{Training Objective.}
The rectified flow connects two arbitrary distributions $y_0\sim\pi_0, y_1\sim \pi_1$. In a manner similar to diffusion models or consistency models, we have $\pi_0=q(y), \pi_1=\mathcal{N}(0, \mathbf{I})$. The forward rectified flow process linearly interpolates $y_t$ as:
\begin{align}
    y_t=t y_1 + (1-t) y_0, t\in[0,1]. \nonumber
\end{align}
We want to learn the velocity at any $t$: $v(y_t, t)=\frac{d}{dt}y_t=y_1-y_0$. It is optimized through the loss:
\begin{align}
    v_1=\arg\min_v\int_0^1 \mathbb{E}_{y_0\sim\pi_0, y_1\sim\pi_1}[||(y_1-y_0)-v(y_t, t)||^2]dt, y_t=t y_1+(1-t)y_0.
    \label{eq:rec_flow1}
\end{align}
By optimizing above Eq.~\eqref{eq:rec_flow1} we derive \texttt{1-Rectified Flow} $v_1$. After ideal optimization process, $v_1$ already achieves certain deterministic mapping for each pair of sample $y_0\sim\pi_0$ and $y_1\sim\pi_1$. We use it generate paired samples $(y_0, y_1)$ for further rectifying it to get \texttt{2-Rectified Flow} $v_2$, which is also called the \textit{Reflow} process:
\begin{align}
    v_2=\arg\min_v\int_0^1 \mathbb{E}_{y_0\sim\pi_0, y_1=\texttt{Flow}_1(y_0)}[||(y_1-y_0)-v(y_t, t)||^2]dt, y_t=t y_1+(1-t)y_0.
\label{eq:rec_flow2}
\end{align}
where $\texttt{Flow}_k(y_0)=y_0+\int_0^1v_k(y_t, t) dt$. This helps to find more straighter and shorter path for connecting paired $(y_0, y_1)$ than \texttt{1-Rectified Flow} $v_1$. Iteratively, it can achieve $k$\texttt{-Rectified Flow} with $O(1/k)$ rate of straightness of the mappings.

\paragraph{Distillation.}
The above process can also be viewed as a distillation process which helps to find straighter paths and reduces sampling steps from a pre-trained RF model.
The optimization objective for RF distillation is:
\begin{align}
    v=\arg\min_v \mathbb{E}_{y_0\sim\pi_0}[d(\texttt{Flow}_k(y_0), y_0+v(y_0))] \nonumber
\end{align}
with $d(\cdot, \cdot)$ is a distance metric. This objective is actually the same as $k$\texttt{-Rectified Flow}, with $k$ being the number of iterations of \textit{Reflow} for the distillation. This is for one-step distillation as $y_0+v(y_0)$ is one-step inference.

\subsubsection{Inference}

As $\pi_0$ and $\pi_1$ are symmetric in above formulation, the inference process of rectified flow can be bidirectional: $\pi_0\rightarrow \pi_1$ or $\pi_1\rightarrow \pi_0$, with a simple sign transformation of velocity. For $\pi_0\rightarrow \pi_1$, the process is straightforward by $\texttt{Flow}_k(y_0)=y_0+\int_0^1v_k(y_t, t) dt, y_0\sim \pi_0$. For discrete timesteps $[t_0, t_1,\dots, t_N]$, $\hat{y}_{t_0}=y_{t_N}+\sum_{i=1}^N (t_i-t_{i-1})v_k(y_{t_i}, t_i)$ with $x_{t_N}\sim\mathcal{N}(0, \mathbf{I})$. On the contrary, for $\pi_1\rightarrow \pi_0$, which is in consistency with the practical requirement of mapping noise back to the original data distribution, this can be achieved by redefining $v=-v$ while utilizing the same sampling formulas. By default, we will use this redefined velocity $v=y_0-y_1$ as the mapping function from noise $y_1$ to data sample $y_0$ in subsequent sections to ensure consistency.

\subsubsection{InstaFlow-prediction}
\begin{figure*}[htbp]
    \centering
\includegraphics[width=0.3\textwidth]{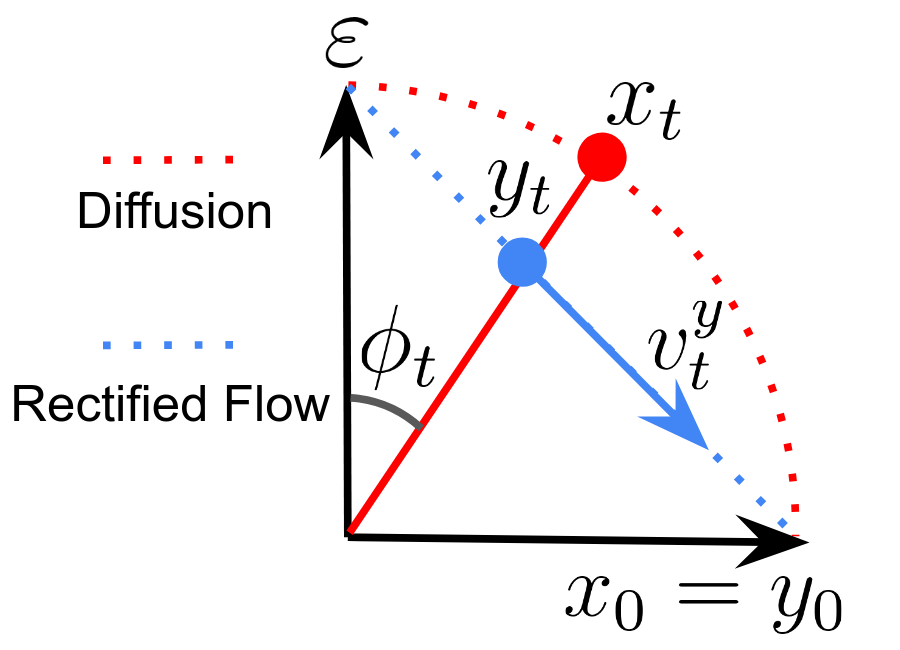}
    \caption{Relationship of $v$-prediction in diffusion and InstaFlow-prediction in rectified flow
    }
    \label{fig:df_rf}
\end{figure*}

The InstaFlow-prediction~\cite{liu2023instaflow} follows the previously introduced rectified flow (RF) method.
In InstaFlow-prediction, the method works on a conjugate space $(y_t, v_\theta^y)_{t=0}^T\in (\mathcal{Y}\times \mathcal{V})^T$ instead of original $(x_t, \epsilon_\theta)_{t=0}^T\in(\mathcal{X}\times \mathcal{E})^T$, with a one-to-one transformation for each $x_t$ and $y_t$, $\forall t\in[T]$, and $y_0=x_0$, as visualized in Fig.~\ref{fig:df_rf}. We will use the upperscript to distinguish between velocity $v^x$ along the diffusion model trajectory and velocity $v^y$ along the RF trajectory.

The InstaFlow network (following RF) directly predicts the target velocity $\tilde{v}^y$ along the RF trajectory, as the difference of the clean sample and Gaussian noise:
\begin{align}
    v_\theta^y\approx \tilde{v}^y = x_0 - \varepsilon
    \label{eq:rf_vel}
\end{align}
Since the RF sample $y_t$ is a scaled version of diffusion sample $x_t$ as:
\begin{align}
y_t=\frac{x_t}{\sqrt{\bar{\alpha}_t}+\sqrt{1-\bar{\alpha}_t}}=\gamma_t x_0 + (1-\gamma_t)\varepsilon, \gamma_t=\frac{\sqrt{\bar{\alpha}_t}}{\sqrt{\bar{\alpha}_t}+\sqrt{1-\bar{\alpha}_t}},
\label{eq:rf_formula}
\end{align}
which satisfies $y_0=x_0$.

\paragraph{$x_\theta$-prediction in InstaFlow.}
Given the InstaFlow network output $v_\theta^y$, we can derive the prediction of original sample $x_\theta$ as:
\begin{align}
    \gamma_t x_\theta &= y_t - (1-\gamma_t)(x_\theta - v_\theta^y)\nonumber\\
    x_\theta &= y_t + (1-\gamma_t)v_\theta^y = y_t + \frac{\sqrt{1-\bar{\alpha}_t}}{\sqrt{\bar{\alpha}_t}+\sqrt{1-\bar{\alpha}_t}}v_\theta^y
    \label{eq:insta_x0}\\
    x_\theta&=\frac{(\sqrt{\bar{\alpha}_t}+\sqrt{1-\bar{\alpha}_t})y_t-\sqrt{1-\bar{\alpha}_t}\epsilon_\theta}{\sqrt{\bar{\alpha}_t}}\nonumber
\end{align}
by replacing $x_0$ with $x_\theta$ in Eq.~\eqref{eq:rf_formula}.
\paragraph{$\epsilon_\theta$-prediction in InstaFlow.}
Given the InstaFlow network output $v_\theta^y$, the prediction of noise $\epsilon_\theta$ as:
\begin{align}
    (1-\gamma_t)\epsilon_\theta = y_t - \gamma_t (v_\theta^y + \epsilon_\theta)\nonumber\\
    \epsilon_\theta = y_t - \gamma_t v_\theta^y = y_t - \frac{\sqrt{\bar{\alpha}_t}}{\sqrt{\bar{\alpha}_t}+\sqrt{1-\bar{\alpha}_t}}v_\theta^y
    \label{eq:v_eps_transform_instaflow}
\end{align}
by replacing $\varepsilon$ with $\epsilon_\theta$ in Eq.~\eqref{eq:rf_formula}.

\paragraph{Score estimation in InstaFlow.}
The score estimation follows:
\begin{align*}
    s_\theta(x_t, t) = -\frac{x_t-\sqrt{\bar{\alpha}_t}x_\theta}{1-\bar{\alpha}_t}, \quad x_t=y_t\cdot (\sqrt{\bar{\alpha}_t}+\sqrt{1-\bar{\alpha}_t})
\end{align*}
with the $x_t\rightarrow y_t$ transformation caused by the InstaFlow prediction. $y_t$ follows Eq.~\eqref{eq:rf_formula} and $x_\theta$ follows Eq.~\eqref{eq:insta_x0}.

\paragraph{Posterior distribution of $x_{t-1}$ in InstaFlow.}

Original Eq.~\eqref{eq:x_t_1_2} for $x_{t-1}$ prediction now becomes:
\begin{align*}
    x_{t-1}(x_t, x_0, \varepsilon)
    &=\frac{\sqrt{\bar{\alpha}_{t-1}}(1-\alpha_t)}{1-\bar{\alpha}_t}x_\theta+\frac{\sqrt{\alpha_t}(1-\bar{\alpha}_{t-1})}{1-\bar{\alpha}_t}x_t+\sigma_t \varepsilon\\
    \Leftrightarrow y_{t-1}\beta_{t-1} &= \frac{\sqrt{\bar{\alpha}_{t-1}}(1-\alpha_t)}{1-\bar{\alpha}_t} ( y_t + (1-\gamma_t)v_\theta^y)+\frac{\sqrt{\alpha_t}(1-\bar{\alpha}_{t-1})}{1-\bar{\alpha}_t}y_t \beta_t+\sigma_t \varepsilon  
\end{align*}
where $\beta_t=\sqrt{\bar{\alpha}_t}+\sqrt{1-\bar{\alpha}_t}, \sigma_t=\sqrt{\frac{(1-\alpha_t)(1-\bar{\alpha}_{t-1})}{1-\bar{\alpha}_t}}$.

Detailed process of InstaFlow inference with $v$-prediction is shown in Alg.~\ref{alg:v_if}.

\begin{algorithm}[H]
\caption{InstaFlow inference with $v$-prediction}
\begin{algorithmic}[1]
\State $y_T\sim\mathcal{N}(0, \mathbf{I})$
\For{$t \in [T]$}
    \State predict velocity $v_\theta^y(y_t, t)$
    \State $\hat{x}_0 \gets y_t + \frac{\sqrt{1-\bar{\alpha}_t}}{\sqrt{\bar{\alpha}_t}+\sqrt{1-\bar{\alpha}_t}}v_\theta^y$ by Eq.~\eqref{eq:insta_x0}
    \State $\mu_\theta(y_{t-1})\gets [\frac{(1 - \alpha_t)\sqrt{\bar{\alpha}_t}}{1-\bar{\alpha}_t}\hat{x}_0  + \frac{\sqrt{\alpha}_t(1-\bar{\alpha}_{t-1})}{1-\bar{\alpha}_t} y_t\beta_t]/\beta_{t-1}$ as Eq.~\eqref{eq:x_t_1_2}
    \State $\sigma_t \gets \sqrt{(1 - \alpha_t)\frac{1-\bar{\alpha}_{t-1}}{1-\bar{\alpha}_t}}/\beta_{t-1}$
    \State $y_{t-1} \gets \mu_\theta + \sigma_t \cdot \varepsilon,  \varepsilon \sim \mathcal{N}(0,\mathbf{I})$
\EndFor
\end{algorithmic}
\label{alg:v_if}
\end{algorithm}

Practically, there exists a simpler sampling method for InstaFlow inference as Alg.~\ref{alg:v_if_simple}, which no longer requires estimation of posterior mean and variance. The disadvantage is that the simpler form cannot be used to estimate the probability of samples due do lack of mean and standard deviation for the distribution.

\begin{algorithm}[H]
\caption{InstaFlow inference with $v$-prediction (simplified)}
\begin{algorithmic}[1]
\State $y_T\sim\mathcal{N}(0, \mathbf{I})$
\For{$t \in [T]$}
    \State predict velocity $v_\theta^y(y_t, t)$
    \State $\hat{x}_0 \gets y_t + \frac{\sqrt{1-\bar{\alpha}_t}}{\sqrt{\bar{\alpha}_t}+\sqrt{1-\bar{\alpha}_t}}v_\theta^y$ by Eq.~\eqref{eq:insta_x0}
    \State $y_{t-1} \gets (\sqrt{\bar{\alpha}_{t-1}}\hat{x}_0 + \sqrt{1-\bar{\alpha}_{t-1}} \varepsilon)/\beta_{t-1},  \varepsilon \sim \mathcal{N}(0,\mathbf{I})$
\EndFor
\end{algorithmic}
\label{alg:v_if_simple}
\end{algorithm}

\paragraph{InstaFlow prediction with DDIM sampling.} What if we use both InstaFlow $v$-prediction and also DDIM sampling? The inference method will be different from Alg.~\ref{alg:v_if}.

Specifically, as discussed previously in Sec.~\ref{sec:diff_inf}, Eq.~\eqref{eq:x_t_1_2} cannot be applied for DDIM sampling on subsequence. Instead, Eq.~\eqref{eq:x_t_1_3_general} should be used in DDIM sampling, with certain changes considering InstaFlow prediction:
\begin{align}
\mu(x_{t_{n-1}})&=\sqrt{\bar{\alpha}_{t_{n-1}}}\hat{x}_0 + \sqrt{1-\bar{\alpha}_{t_{n-1}} -\sigma_{t_n}^2}\epsilon_\theta\nonumber\\
\beta_{t_{n-1}} \mu(y_{t_{n-1}})&=\sqrt{\bar{\alpha}_{t_{n-1}}}\hat{x}_0 + \sqrt{1-\bar{\alpha}_{t_{n-1}} -\sigma_{t_n}^2}(y_{t_n} - \frac{\sqrt{\bar{\alpha}_{t_n}}}{\beta_{t_n}}v_\theta^y) \quad \text{(by Eq.~\eqref{eq:v_eps_transform_instaflow})}
\label{eq:insta_ddim_x_t_1}
\end{align}
This provides algorithm for InstaFlow prediction with DDIM sampling for model inference, as Alg.~\ref{alg:v_if_ddim}.

\begin{algorithm}[H]
\caption{InstaFlow inference with $v$-prediction and DDIM sampling}
\begin{algorithmic}[1]
\State $y_{t_N}\sim\mathcal{N}(0, \mathbf{I}), \{t_n\}_{n=1}^N\subseteq[T]$
\For{$n\in [N]$}
    \State predict velocity $v_\theta^y(y_{t_n}, t_n)$
    \State $\hat{x}_0 \gets y_{t_n} + \frac{\sqrt{1-\bar{\alpha}_{t_n}}}{\sqrt{\bar{\alpha}_{t_n}}+\sqrt{1-\bar{\alpha}_{t_n}}}v_\theta^y$ by Eq.~\eqref{eq:insta_x0}
    \State $\mu_\theta(y_{t_{n-1}})\gets [\sqrt{\bar{\alpha}_{t_{n-1}}}\hat{x}_0 + \sqrt{1-\bar{\alpha}_{t_{n-1}} -\sigma_{t_n}^2}(y_{t_n} - \frac{\sqrt{\bar{\alpha}_{t_n}}}{\beta_{t_n}}v_\theta^y)]/\beta_{t_{n-1}}$ as Eq.~\eqref{eq:insta_ddim_x_t_1}
    \State $\sigma_{t_n} \gets \sqrt{(1 - \alpha_{t_n})\frac{1-\bar{\alpha}_{t_{n-1}}}{1-\bar{\alpha}_{t_n}}}/\beta_{t_{n-1}}$
    \State $y_{t_{n-1}} \gets \mu_\theta + \sigma_{t_n} \cdot \varepsilon,  \varepsilon \sim \mathcal{N}(0,\mathbf{I})$
\EndFor
\end{algorithmic}
\label{alg:v_if_ddim}
\end{algorithm}

\subsubsection{Distillation}
In this section, we explore the possibilities of applying the rectified flow (RF) training objective in the distillation process from a diffusion model, different from the distillation from RF model itself in Sec.~\ref{subsec:rf_train_infer}.

To leverage the loss of rectified flow in distillation from a diffusion model, there are two potential approaches: 1. learning a separate RF network for velocity prediction on RF trajectories; 2. reparameterize RF $v$-prediction with diffusion $\epsilon$-prediction.

\paragraph{Approach 1.}

Take the pre-trained $\epsilon$-prediction model $\epsilon_\phi$ to generate paired samples $(\varepsilon, \hat{x}_0)$, where $\hat{x}_0=\text{Denoise}^N(\varepsilon;\epsilon_\phi)=\text{Denoise}(\text{Denoise}...\text{Denoise}(\varepsilon;\epsilon_\phi))$ as defined by Eq.~\eqref{eq:multi_step_denoise} given teacher model denoising function with $N$-step DDIM sampling schedule. Then take the training loss of \texttt{2-Rectified Flow} as Eq.~\eqref{eq:rec_flow2} to learn an individual $v(\theta)$, which is not reparameterized from $\epsilon_\theta$ by $v$-prediction. The distillation process follows the loss:
\begin{align}
    \min_\theta\int_0^1 \mathbb{E}_{\varepsilon\sim\mathcal{N}(0, \mathbf{I}), \hat{x}_0=\text{Denoise}^N(\varepsilon;\epsilon_\phi)}[||(\hat{x}_0-\varepsilon)-v_\theta^y(x_t, t)||^2]dt,
\label{eq:rec_flow_idea1}
\end{align}
with $x_t=t x_0 + (1-t) \varepsilon$. This essentially treats the pre-trained diffusion model with DDIM scheduler as \texttt{ 1-rectified flow}. 

\paragraph{Approach 2.}
It can be ill-posed to directly enforce rectified flow loss for student model $\epsilon_\theta$ (reparameterize it into $v_\theta^x$) in the distillation process:
\begin{align}
    \min_\theta\int_0^1 \mathbb{E}_{\varepsilon\sim\mathcal{N}(0, \mathbf{I}), \hat{x}_0=\text{Denoise}^N(\varepsilon;\epsilon_\phi)}[||(x_0-\varepsilon)-v_\theta^x(t)||^2]dt,
\label{eq:rec_flow_idea}
\end{align}
with $v_\theta^x(t)=\sqrt{1-\bar{\alpha}_t}x_\theta(x_{t'}, t') - \sqrt{\bar{\alpha}_t}\varepsilon$ representing $v$-prediction/parameterization in diffusion and $x_{t'}=\sqrt{\bar{\alpha}_{t'}}x_0+\sqrt{1-\bar{\alpha}_{t'}}\varepsilon$, it becomes evident that DDIM/DDPM flows follow a circular trajectory with a constant norm, whereas RF flows along a straight line with a varying norm, as illustrated in Fig.~\ref{fig:df_to_rf}. As a result, the velocity cannot be directly matched. Here, $v_\theta^x(t)$ refers to the $v$-prediction of the diffusion model, which characterizes the velocity along the diffusion trajectory, rather than the $v$-prediction $v_\theta^y(t)$ along the RF trajectory. Given the difference in noise schedules, a mapping is required to align the velocity from the diffusion trajectory with the RF trajectory.

\subsubsection{Velocity Mapping}
\label{sec:vel_diff_rf}
\begin{figure}[htbp]
    \centering
\includegraphics[width=0.4\textwidth]{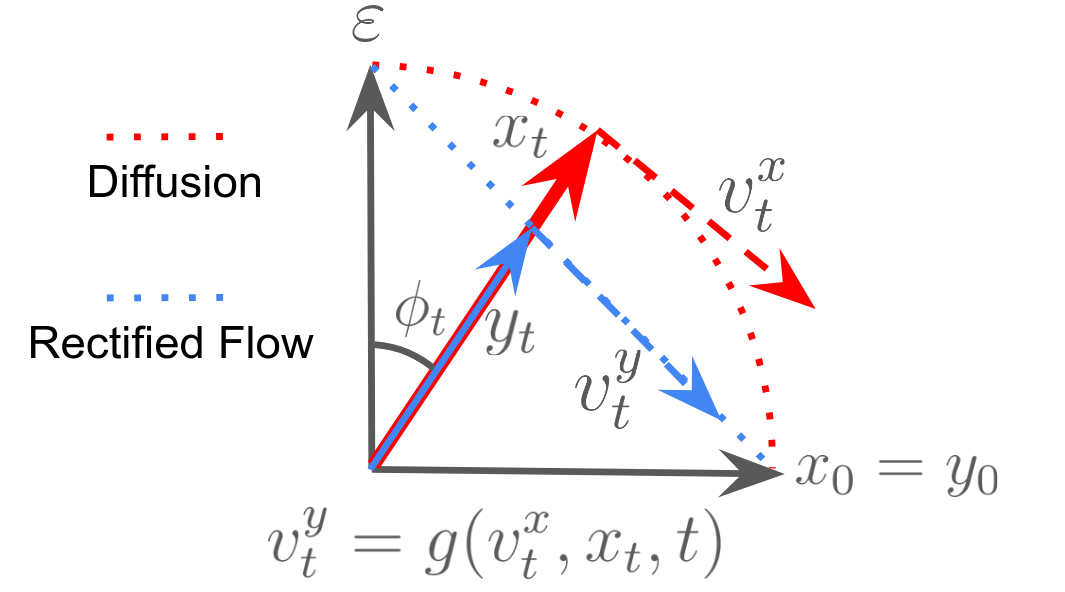}
    \caption{Velocity mapping from $v$-prediction in diffusion to InstaFlow-prediction in rectified flow
    }
    \label{fig:df_to_rf}
\end{figure}

We emphasize the relationship between velocities on the diffusion trajectory and the rectified flow trajectory.
To correctly apply above Approach 2 for RF, we need to map the velocity on the diffusion trajectory $v_\theta^x(t)$ with the velocity on the RF trajectory $v_\theta^y(t)$, as demonstrated in Fig.~\ref{fig:df_to_rf} for a vector representation of the variables.

In Sec.~\ref{sec:v_in_rf}, we prove that there exists a velocity mapping between rectified flow velocity $v^y_t$ and diffusion velocity $v^x_t$:
\begin{align}
    v^y_t = \frac{v^x_t b-x_t(\sqrt{1-\bar{\alpha}_t}-\sqrt{\bar{\alpha}_t})}{b^2}\nonumber
\end{align}
also 
\begin{align}
    v^y_t =\frac{x_0-\varepsilon}{1+2\sqrt{\bar{\alpha}_t}\sqrt{1-\bar{\alpha}_t}}=(x_0-\varepsilon)\frac{||y_t||^2}{||x_t||^2}\nonumber
\end{align}
which can be leveraged to form the rectified loss:
\begin{align}
    v_2=\min_v\int_0^1 \mathbb{E}_{\varepsilon\sim\mathcal{N}(0, \mathbf{I}), x_0=\text{Denoise}(\varepsilon)}[||(1+2\sqrt{\bar{\alpha}_t}\sqrt{1-\bar{\alpha}_t})v_\theta^y(t)-(x_0-\varepsilon)||^2]dt, \nonumber
\end{align}
with,
\begin{align*}
    v_\theta^y(t) &= \frac{v^x_t b_t-x_t(\sqrt{1-\bar{\alpha}_t}-\sqrt{\bar{\alpha}_t})}{b_t^2}\\
    &=\frac{(\sqrt{1-\bar{\alpha}_t}x_\theta(x_t, t)-\sqrt{\bar{\alpha}_t}\varepsilon) b_t-x_t(\sqrt{1-\bar{\alpha}_t}-\sqrt{\bar{\alpha}_t})}{b_t^2}\\
    (1+2\sqrt{\bar{\alpha}_t}\sqrt{1-\bar{\alpha}_t})v_\theta^y(t)-(x_0-\varepsilon)&=[\sqrt{\bar{\alpha}_t(1-\bar{\alpha}_t)}+(1-\bar{\alpha}_t)](x_\theta - x_0)\\
    &=[(1-\bar{\alpha}_t)(\sqrt{\bar{\alpha}_t}+\sqrt{1-\bar{\alpha}_t})](\varepsilon - \epsilon_\theta)
\end{align*}
with $x_\theta=\frac{1}{\sqrt{\bar{\alpha}_t}}x_t-\frac{\sqrt{1-\bar{\alpha}_t}}{\sqrt{\bar{\alpha}_t}}\epsilon_\theta$, which is essentially a weighted regression or diffusion loss. This loss can be utilized to directly fine-tune the original diffusion model $\epsilon_\theta$ (via $x_\theta$) by enforcing the rectified flow loss on its mapped velocity, transitioning from the diffusion trajectory to the RF trajectory.

After training, the method additionally supports RF-based sampling. By transforming the diffusion velocity into RF velocity, it enables direct sampling along a straight RF trajectory, thereby facilitating efficient and effective few-step sampling. 

The above discussion offers a unified perspective on the diffusion model and the RF model, integrating both their training objectives and inference processes through the mapping of velocities between the diffusion and RF processes. The next section will introduce TrigFlow, a novel model formulation designed to unify the consistency model and the RF model.

\subsection{TrigFlow}
TrigFlow~\cite{lu2024simplifying} modifies the trigonometric interpolant~\cite{albergo2022building} to unify continuous-time consistency models and rectified flow with $v$ prediction. 
\subsubsection{Training}
\label{sec:trigflow_train}
Recall in Sec.~\ref{sec:consist_train}, the consistency model follows EDM parameterization~\cite{karras2022elucidating} to satisfy the boundary condition ($f_\theta(x_\epsilon, \epsilon)=x_\epsilon\approx x_0$ when $\epsilon \rightarrow 0$):
\begin{align}
f_\theta(x_t,t) &= c_{\text{skip}}(t) x_t + c_{\text{out}}(t)x_\theta(c_{\text{in}}(t)x_t, c_{\text{noise}}(t), c)
\label{eq:cm_param}
\end{align}
where $c_{\text{skip}}(t)=\frac{\sigma^2}{(t-\epsilon)^2+\sigma^2}$, $c_{\text{out}}(t)=\frac{\sigma(t-\epsilon)}{\sqrt{t^2+\sigma^2}}$, $c_\text{in}(t)=\frac{1}{\sqrt{
t^2+\sigma^2}}$ and $c_\text{noise}(t)$ is a transformation of $t$ for time conditioning, $\sigma$ is the standard deviation of training sample distribution. TrigFlow takes a different set of coefficients with the boundary condition also satisfied:
\begin{align}
f_\theta(x_t, t) = \cos(t)x_t - \sin(t)\sigma v_\theta(\frac{x_t}{\sigma}, c_\text{noise}(t), c)
\label{eq:trigflow_param}
\end{align}
where $v_\theta$ is the normalized (by $\sigma$) velocity prediction with target $v_t$.
TrigFlow can be trained using either the diffusion objective or the consistency training objective, effectively unifying the principles of rectified flow and consistency models. When employing the diffusion training objective, TrigFlow adopts velocity prediction, akin to rectified flow:
\begin{align}
\mathcal{L}_\text{TF-diffusion}=\mathbb{E}_{x_0, \varepsilon, t}[||\sigma v_\theta(\frac{x_t}{\sigma}, c_\text{noise}(t), c) - v_t||_2^2]\nonumber
\end{align}
$v_t=\cos(t)\varepsilon - \sin(t)x_0$ is the velocity for diffusion process $x_t=\cos(t)x_0+\sin(t)\varepsilon, t\in[0, \frac{\pi}{2}], \varepsilon\sim\mathcal{N}(0, \sigma^2\mathbf{I})$. The angular parameterization of diffusion process will be discussed later in Sec.~\ref{sec:v_diff}. The choice of coefficients $c_\text{skip}$, $c_\text{out}$ and $c_\text{in}$ ensures the input of $v_\theta$ and its target to have unit variance, for the convenience of training. 

Recall the discrete-time consistency training (CT) objective defined in Eq.~\eqref{eq:cm_sample}. In the continuous-time setting, as $\Delta t \to 0$, its gradient is proven to converge to the following form:
\begin{align}
\nabla_\theta\mathcal{L}_\text{TF-CT}=\nabla_\theta \mathbb{E}_{x_t, t}[\lambda(t)f^\intercal_\theta(x_t, t)\frac{d f_{\theta^-}(x_t,t)}{dt}]\nonumber
\end{align}
for $l_2$ distance metric $d(x,y)=||x-y||_2^2$.

With Eq.~\eqref{eq:trigflow_param} plugged in, we have the gradients:
\begin{align}
\nabla_\theta \mathbb{E}_{x_t, t}[-\lambda(t)\sin(t)\sigma v^\intercal_\theta(\frac{x_t}{\sigma}, c_\text{noise}(t), c)\frac{d f_{\theta^-}(x_t,t)}{dt}]\nonumber
\end{align}
as $\cos(t)x_t$ in $f_\theta(x_t, t)$ does not depend on $\theta$.

Additional techniques are proposed to improve the training with this objective gradient in original paper~\cite{lu2024simplifying}.

\subsubsection{Inference}
After training, TrigFlow model can be sampled with DDIM schedule as:
\begin{align}
x_t=\cos(s-t)x_s - \sin(s-t)\sigma v_\theta(\frac{x_s}{\sigma}, c_\text{noise}(s), c)\nonumber
\end{align}
for a single step denoising from timestep $s$ to $t$ ($s>t$), starting from Gaussian noise $x_{T}\sim\mathcal{N}(0, \mathbf{I}), T=\frac{\pi}{2}$.

\subsection{Prediction Parameterization}
\label{sec:parameterization}

\begin{figure*}[htbp]
    \centering
    \quad\quad\quad\quad\quad
\includegraphics[width=0.5\textwidth]{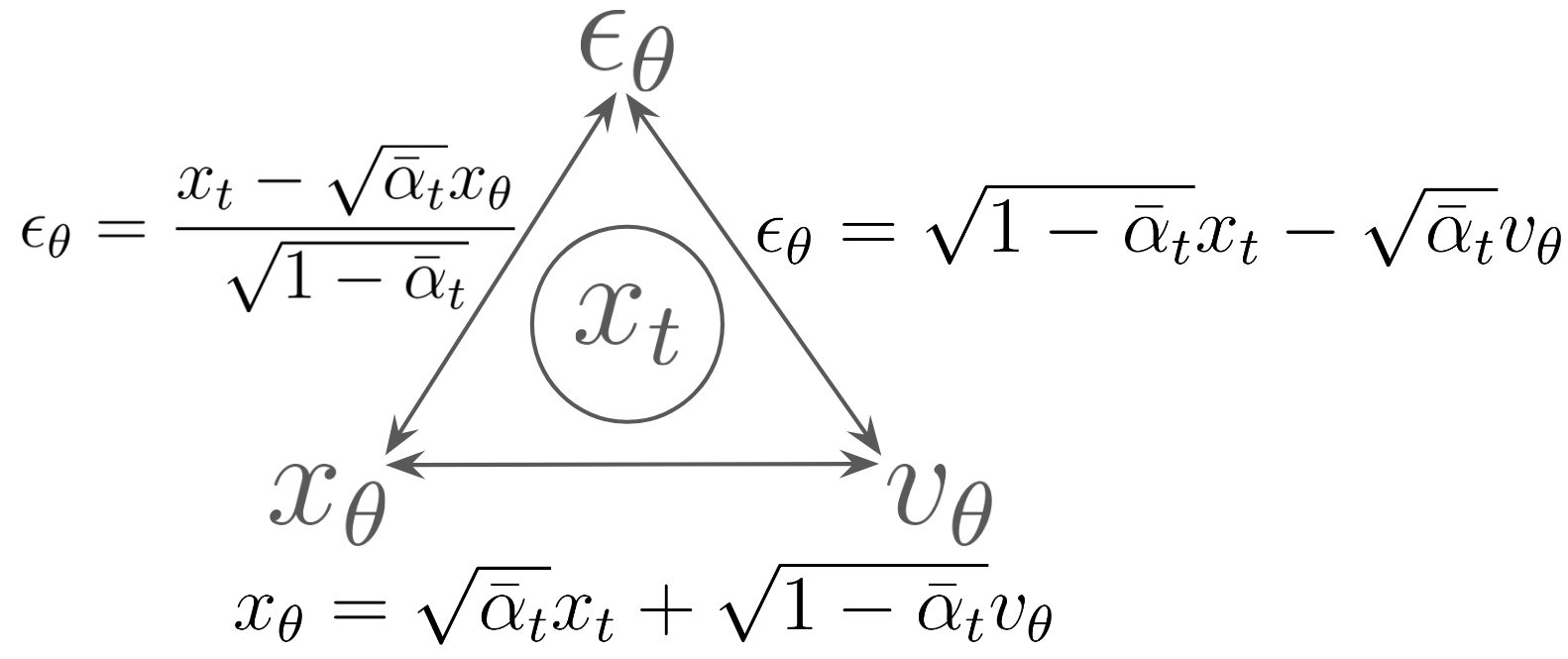}
    \caption{The ``triangular''-formula in diffusion model for three parameterization: $\epsilon$-prediction, $x$-prediction and $v$-prediction. 
    }
    \label{fig:df_triangle}
\end{figure*}

\begin{figure*}[htbp]
    \centering
\includegraphics[width=0.7\textwidth]{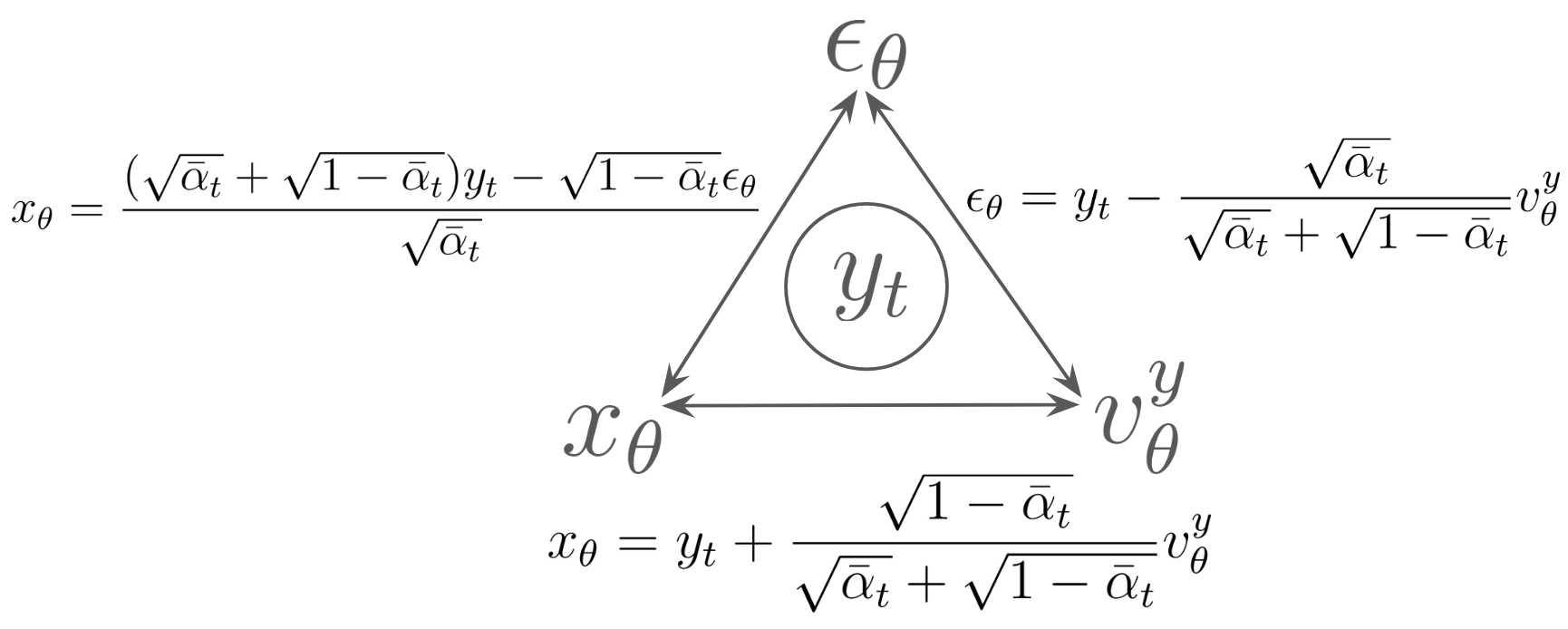}
    \caption{The ``triangular''-formula in rectified flow model for three parameterization: $\epsilon$-prediction, $x$-prediction and $v$-prediction. 
    }
    \label{fig:rf_triangle}
\end{figure*}
In this section, we discuss different prediction parameterizations for various models, including diffusion model, consistency model and rectified flow. The relationship among different parameterizations connect these models, which is useful for practical transformation from one type of model to another.

For diffusion model, we have the ``triangular''-formula for three parameterization: $\epsilon$-prediction, $x$-prediction and $v$-prediction, as Fig.~\ref{fig:df_triangle}.

For rectified flow model, we have the ``triangular''-formula for three parameterization: $\epsilon$-prediction, $x$-prediction and $v$-prediction, as Fig.~\ref{fig:rf_triangle}.

\subsubsection{$\epsilon$-prediction and $x$-prediction in Diffusion Model}
According to Tweedies's formula, the $\epsilon$-prediction and $x$-prediction can be mutually transformed via:
\begin{align*}
\epsilon_\theta&=\frac{x_t-\sqrt{\bar{\alpha}_t}x_\theta}{\sigma_t}\\
x_\theta&=\frac{x_t-\sigma_t \epsilon_\theta}{\sqrt{\bar{\alpha}_t}}
\end{align*}
with $\sigma_t=\sqrt{1-\bar{\alpha}_t}$.

\begin{figure*}[htbp]
    \centering
\includegraphics[width=0.99\textwidth]{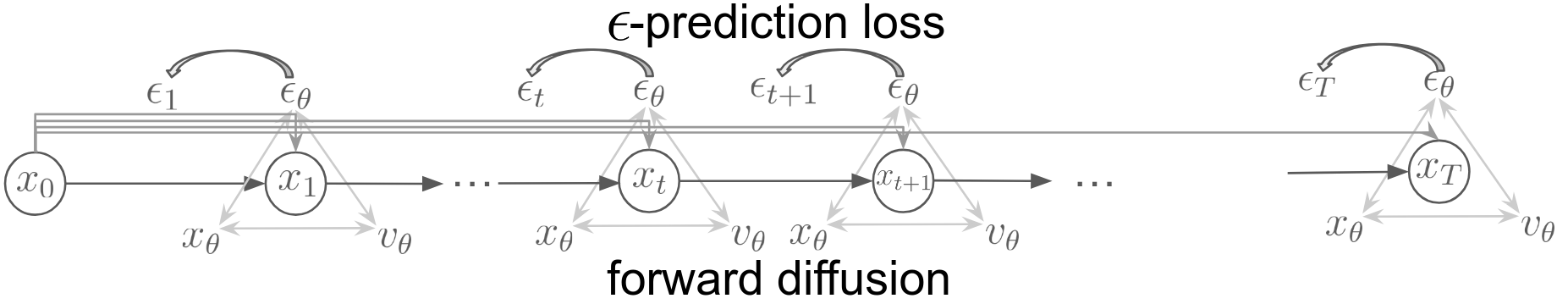}
    \caption{$\epsilon$-prediction loss along the diffusion trajectory.
    }
    \label{fig:diff_eps_loss}
\end{figure*}

\begin{figure*}[htbp]
    \centering
\includegraphics[width=0.99\textwidth]{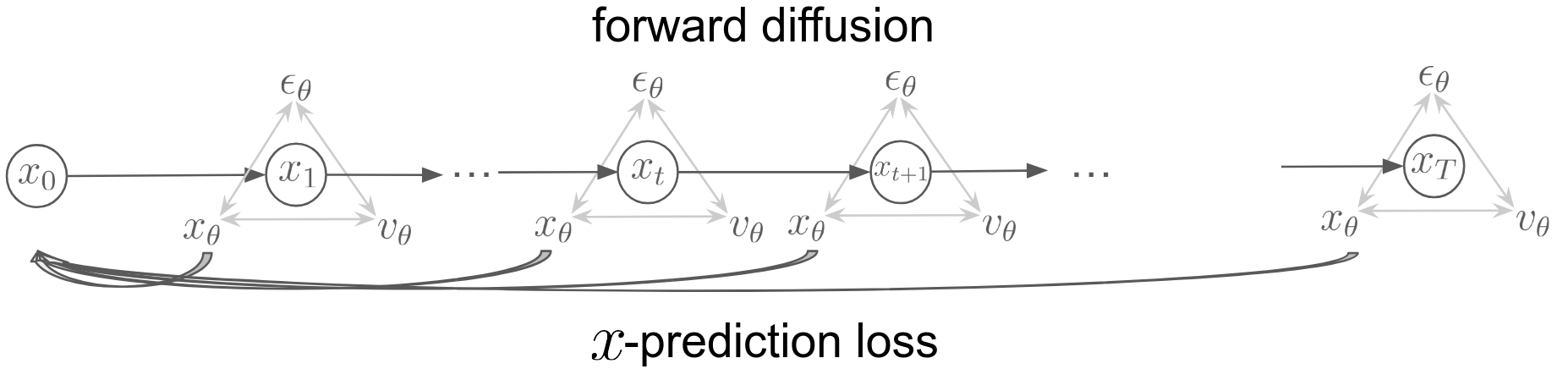}
    \caption{$x$-prediction loss along the diffusion trajectory.
    }
    \label{fig:diff_x_loss}
\end{figure*}

Based on the ``triangular''-formula relationship, we show a diagram of $\epsilon$-prediction loss and $x$-prediction loss for diffusion model as in Fig.~\ref{fig:diff_eps_loss} and Fig.~\ref{fig:diff_x_loss}.

\subsubsection{$v$-prediction in Diffusion Model}
\label{sec:v_diff}
\begin{figure*}[htbp]
    \centering
\includegraphics[width=0.2\textwidth]{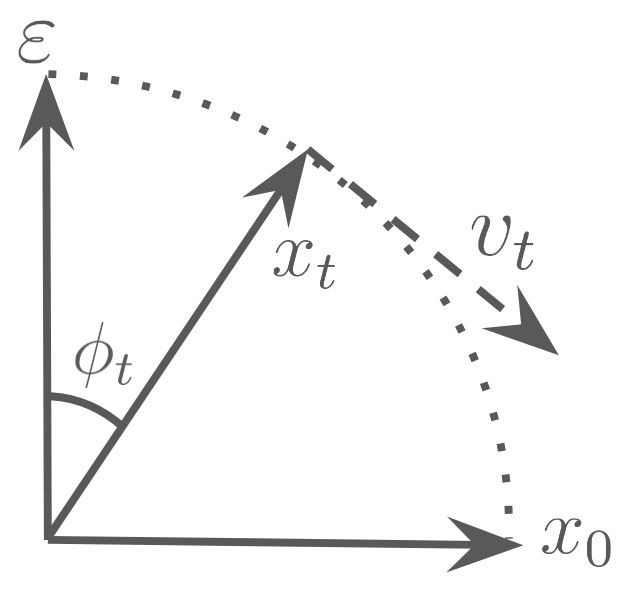}
\hspace{0.2\textwidth}
\includegraphics[width=0.2\textwidth]{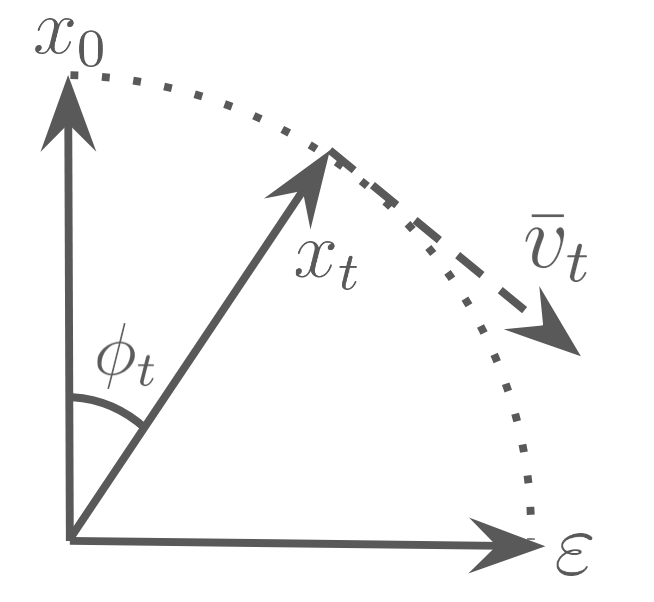}
\caption{Visualization of $v$-prediction (left) and reverse $\bar{v}$-prediction (right) in diffusion process.
    }
    \label{fig:eps_v_ours}
\end{figure*}
Apart from $\epsilon$-prediction, there is also a commonly used $v$-prediction. Eq.~\eqref{eq:xt_x0} can be rewritten in angular coordinate as:
\begin{align}
    x_t = x_0\sin\phi_t  + \varepsilon\cos\phi_t \nonumber
\end{align}
where $\sin\phi_t=\sqrt{\bar{\alpha}_t}, \cos\phi_t=\sqrt{1-\bar{\alpha}_t}$. 
Then angular velocity $v_t=\frac{d x_t}{dt}=\cos\phi_t x_0-\sin\phi_t \varepsilon$ for  reverse diffusion. Velocity for forward diffusion is just $\bar{v}_t=-\frac{d x_t}{dt}=-\cos\phi x_0 + \sin\phi \varepsilon=-\sqrt{1-\bar{\alpha}_t}x_0 + \sqrt{\bar{\alpha}_t}\varepsilon$. Therefore the $v$-prediction (for reverse diffusion by default) is:
\begin{align}
   v_\theta=\sqrt{1-\bar{\alpha}_t}x_\theta - \sqrt{\bar{\alpha}_t}\epsilon_\theta \nonumber
\end{align}
and we have its connection to $\epsilon$- and $x$-prediction:
\begin{align*}
x_\theta &= \sqrt{\bar{\alpha}_t}x_t + \sqrt{1-\bar{\alpha}_t}v_\theta\\
\epsilon_\theta &= \sqrt{1-\bar{\alpha}_t}x_t - \sqrt{\bar{\alpha}_t}v_\theta\\
\Rightarrow \epsilon_\theta&= \frac{x_t}{\sqrt{1-\bar{\alpha}_t}} - \frac{\sqrt{\bar{\alpha}_t}x_\theta}{\sqrt{1-\bar{\alpha}_t}}
\end{align*}

The Fig.~\ref{fig:eps_v_ours} illustrates the $v$-prediction in diffusion process.

Here we intentionally choose the direction of velocity to coincide with the direction of velocity in rectified flow.

Note that practically people sometimes use $v$-prediction for forward diffusion velocity:
$\bar{v}_\theta=\sqrt{\bar{\alpha}_t}\epsilon_\theta - \sqrt{1-\bar{\alpha}_t}x_\theta$, which satisfies:
\begin{align*}
x_\theta &= \sqrt{\bar{\alpha}_t}x_t - \sqrt{1-\bar{\alpha}_t}\bar{v}_\theta\\
\epsilon_\theta &= \sqrt{1-\bar{\alpha}_t}x_t + \sqrt{\bar{\alpha}_t}\bar{v}_\theta\\
\Rightarrow \epsilon_\theta&= \frac{x_t}{\sqrt{1-\bar{\alpha}_t}} - \frac{\sqrt{\bar{\alpha}_t}x_\theta}{\sqrt{1-\bar{\alpha}_t}}
\end{align*}
which corresponds to Fig.~\ref{fig:eps_v_ours} right image. At inference, this just requires to reverse the sign of velocity for sampling from $\varepsilon$ to $x_0$.

\subsubsection{$v$-prediction in Rectified Flow}
\label{sec:v_in_rf}
\begin{figure*}[htbp]
    \centering
\includegraphics[width=0.99\textwidth]{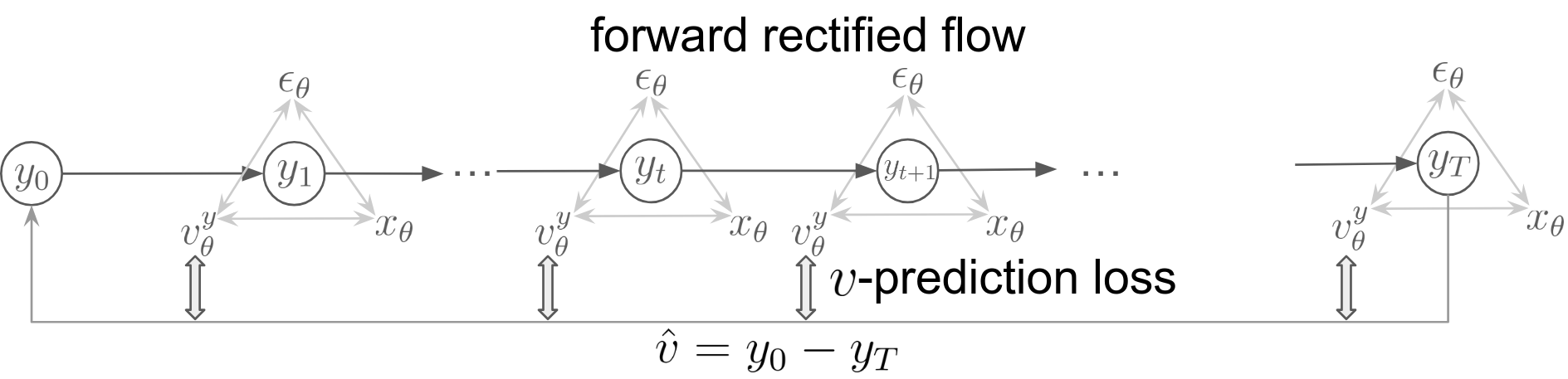}
    \caption{$v$-prediction loss along the RF trajectory.
    }
\label{fig:rf_loss}
\end{figure*}

Notice that the $v$-prediction above (denoted as $v^x_t$) is along the diffusion trajectory, therefore it is different from the RF's velocity prediction (as Eq.~\eqref{eq:rf_vel}) along the RF trajectory denoted as $v^y_t$. 
We also call it InstaFlow-prediction. The diagram of $v$-prediction loss for training the RF model is shown as Fig.~\ref{fig:rf_loss}

Actually, there exists a mapping relationship between the two velocities:
\begin{align}
    v^y_t = \frac{v^x_t b-x_t(\sqrt{1-\bar{\alpha}_t}-\sqrt{\bar{\alpha}_t})}{b^2}, b=\sqrt{\bar{\alpha}_t}+\sqrt{1-\bar{\alpha}_t}\nonumber
\end{align}
with the proof as following.

\textit{Proof}:
Suppose the sample from diffusion model is $x_t$ and sample from rectified flow is $y_t$ (as scaled $x_t$), we have
\begin{align}
    x_t&=\sqrt{\bar{\alpha}_t}x_0 + \sqrt{1-\bar{\alpha}_t}\varepsilon\nonumber \\
    y_t&=\frac{x_t}{\sqrt{\bar{\alpha}_t}+\sqrt{1-\bar{\alpha}_t}}=\gamma_t x_0 + (1-\gamma_t)\varepsilon, \gamma_t=\frac{\sqrt{\bar{\alpha}_t}}{\sqrt{\bar{\alpha}_t}+\sqrt{1-\bar{\alpha}_t}}\nonumber\\
    v^x_t = \dot{x_t}&=\sqrt{1-\bar{\alpha}_t}x_0 - \sqrt{\bar{\alpha}_t}\varepsilon\nonumber\\
    v^y_t=\dot{y_t}&=\frac{a'b-b'a}{b^2}, a=x_t, b=\sqrt{\bar{\alpha}_t}+\sqrt{1-\bar{\alpha}_t}\label{eq:vel_match}\\
    &=\frac{\dot{x_t}b-x_t(\sqrt{1-\bar{\alpha}_t}-\sqrt{\bar{\alpha}_t})}{b^2}\nonumber\\
    &=\frac{(\sqrt{1-\bar{\alpha}_t}x_0-\sqrt{\bar{\alpha}_t}\varepsilon)(\sqrt{\bar{\alpha}_t}+\sqrt{1-\bar{\alpha}_t})-(\sqrt{\bar{\alpha}_t}x_0+\sqrt{1-\bar{\alpha}_t}\varepsilon)(\sqrt{1-\bar{\alpha}_t}-\sqrt{\bar{\alpha}_t})}{1+2\sqrt{\bar{\alpha}_t}\sqrt{1-\bar{\alpha}_t}}\nonumber\\
    &=\frac{x_0-\varepsilon}{1+2\sqrt{\bar{\alpha}_t}\sqrt{1-\bar{\alpha}_t}}\nonumber\\
    &=(x_0-\varepsilon)\frac{||y_t||^2}{||x_t||^2} \nonumber
\end{align}
Eq.~\eqref{eq:vel_match} provides the velocity mapping relationship between rectified flow velocity $v^y_t$ and diffusion velocity $v^x_t$. $\square$

For a clearer intuitive understanding, Fig.~\ref{fig:df_to_rf} illustrates the relationship of $v$-prediction in diffusion and InstaFlow-prediction in rectified flow.

\subsubsection{$\epsilon$-prediction and $x$-prediction in Rectified Flow}

As illustrated in Fig.~\ref{fig:rf_triangle}, there also exist at least three parameterization for RF: 
 $\epsilon$-prediction, $x$-prediction and $v$-prediction. Their relationships follow:
 \begin{align*}
\epsilon_\theta  &= y_t - \frac{\sqrt{\bar{\alpha}_t}}{\sqrt{\bar{\alpha}_t}+\sqrt{1-\bar{\alpha}_t}}v_\theta^y\\
x_\theta&=\frac{(\sqrt{\bar{\alpha}_t}+\sqrt{1-\bar{\alpha}_t})y_t-\sqrt{1-\bar{\alpha}_t}\epsilon_\theta}{\sqrt{\bar{\alpha}_t}}\\
x_\theta &= y_t + \frac{\sqrt{1-\bar{\alpha}_t}}{\sqrt{\bar{\alpha}_t}+\sqrt{1-\bar{\alpha}_t}}v_\theta^y
 \end{align*}
The three equations are mutually consistent.

\subsubsection{$f$-prediction in Consistency Model}

Given diffusion model (DM) $\epsilon_\theta$, using Eq.~\eqref{eq:consist} and \eqref{eq:xt_x0}, consistency model $f_\theta$ can be reparameterized by (proposed by LCM~\cite{luo2023latent}):
\begin{align*}
    f_\theta(x_t,t) &= c_{\text{skip}} x_t + c_{\text{out}}x_\theta(x_t, c, t)\\
    x_\theta&=\frac{x_t-\sqrt{1-\bar{\alpha}_t}\epsilon_\theta(x_t,c,t)}{\sqrt{\bar{\alpha}_t}}
\end{align*}
where $x_\theta$ is the prediction of clean sample $\hat{x}_0$.
It is worth noting that this requires the DM and CM to have aligned timesteps $t$; at least, the DM timesteps need to encompass the CM timesteps.


\begin{figure*}[htbp]
    \centering
\includegraphics[width=0.99\textwidth]{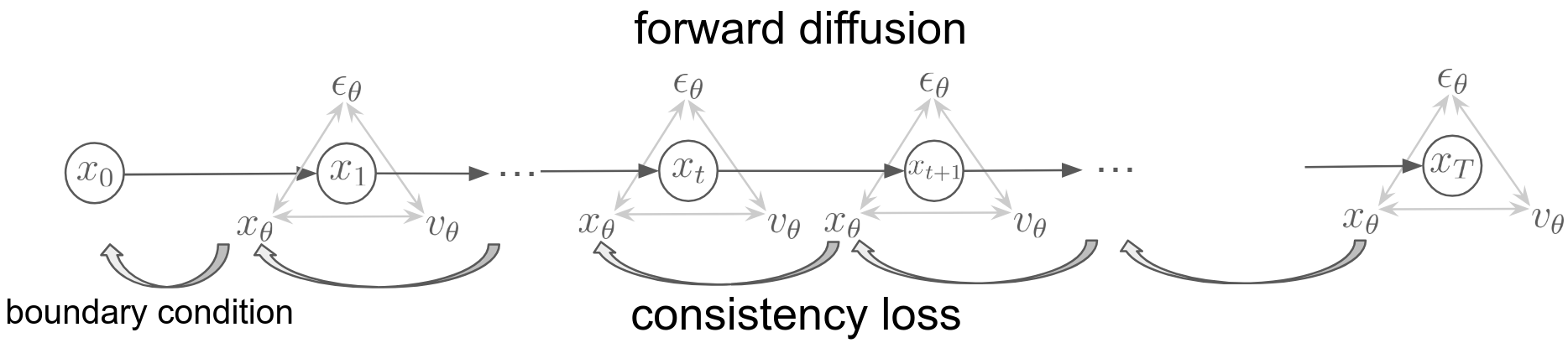}
    \caption{Consistency loss along the diffusion trajectory.
    }
    \label{fig:consist_loss}
\end{figure*}
Based on the ``triangular''-formula relationship, we show a diagram of consistency loss in Fig.~\ref{fig:consist_loss}.

\subsection{Unified Formulation}
For general SDEs connecting two arbitrary distributions $x_0\sim\pi_0, x_1\sim \mathcal{N}(0, \mathbf{I})$, with given sequence of coefficients $\alpha_t, \beta_t, \forall t\in[0,1]$ (corresponding to $\bar{a}_t=\alpha_t, \bar{b}_t=\beta_t, T=1$ in Def.~\ref{def:dm}):
\begin{align}
    x_t=\alpha_t x_0 + \beta_t x_1 \nonumber
\end{align}
We have the following velocity and score relationship:
\begin{align*}
    s(x_t, t)&=-\frac{x_t - \alpha_t x_0}{\beta_t^2}=-\frac{x_1}{\beta_t}  \quad\quad \text{(only for $x_1\sim \mathcal{N}(0, \mathbf{I})$)}\\
    v(x_t, t) &= \frac{\dot{\alpha_t}}{\alpha_t}x_t + \beta_t(\frac{\dot{\alpha_t}}{\alpha_t}\beta_t - \dot{\beta_t}) s(x_t, t)
\end{align*}
with score function $s(x_t,t)=\nabla \log p_t(x_t)$. $\dot{\alpha_t}=\frac{d\alpha_t}{dt}, \dot{\beta}=\frac{d\beta}{dt}$ as time derivatives.

\textit{Proof}:
Given the stochastic process:
\[
x_t = \alpha_t x_0 + \beta_t x_1,
\]
where \(x_0 \sim \pi_0\), \(x_1 \sim \pi_1\), and \(\alpha_t, \beta_t\) are differentiable functions of time \(t \in [0,1]\).

The score function is defined as:
\[
s(x_t, t) = \nabla_{x_t} \log p_t(x_t),
\]
where \(p_t(x_t)\) is the probability density function of \(x_t\).

Since \(x_t\) is a linear transformation of \(x_1\) given \(x_0\), the conditional density is:
\[
p_t(x_t \mid x_0) = \frac{1}{\beta_t} p_{x_1}\left( \frac{x_t - \alpha_t x_0}{\beta_t} \right).
\]

Compute the score function conditioned on \(x_0\):
\[
\begin{aligned}
s(x_t, t) &= \nabla_{x_t} \log p_t(x_t \mid x_0) \\
&= \nabla_{x_t} \log \left[ \frac{1}{\beta_t} p_{x_1}\left( \frac{x_t - \alpha_t x_0}{\beta_t} \right) \right] \\
&= \nabla_{x_t} \left( -\log \beta_t + \log p_{x_1}\left( \frac{x_t - \alpha_t x_0}{\beta_t} \right) \right) \\
&= \frac{1}{\beta_t} \nabla_{x_1} \log p_{x_1}(x_1) \bigg|_{x_1 = \frac{x_t - \alpha_t x_0}{\beta_t}}.
\end{aligned}
\]

Since $x_1 \sim\mathcal{N}(0, \mathbf{I})$, the score function of \(x_1\) is:
\begin{align*}
s_{x_1}(x_1) &= \nabla_{x_1} \left( -\frac{1}{2} (x_1 - \mu)^\top \mathbf{\Sigma}^{-1} (x_1 - \mu) \right)\\
&= -\mathbf{\Sigma}^{-1} (x - \mu)\\
&= -x_1
\end{align*}
with $\mu=0, \mathbf{\Sigma}=\mathbf{I}$.
Therefore:
\[
s(x_t, t) = -\frac{x_t - \alpha_t x_0}{\beta_t^2} = -\frac{x_1}{\beta_t}.
\]

Compute the time derivative of \(x_t\):
\[
\frac{dx_t}{dt} = \dot{\alpha}_t x_0 + \dot{\beta}_t x_1,
\]
where \(\dot{\alpha}_t = \frac{d\alpha_t}{dt}\) and \(\dot{\beta}_t = \frac{d\beta_t}{dt}\).

Express \(x_0\) and \(x_1\) in terms of \(x_t\) and \(s(x_t, t)\):
\[
\begin{aligned}
x_1 &= -\beta_t s(x_t, t), \\
x_0 &= \frac{x_t + \beta_t^2 s(x_t, t)}{\alpha_t}.
\end{aligned}
\]

Substitute back into \(\frac{dx_t}{dt}\):
\[
\begin{aligned}
\frac{dx_t}{dt} &= \dot{\alpha}_t \left( \frac{x_t + \beta_t^2 s(x_t, t)}{\alpha_t} \right) + \dot{\beta}_t \left( -\beta_t s(x_t, t) \right) \\
&= \frac{\dot{\alpha}_t}{\alpha_t} x_t + \dot{\alpha}_t \left( \frac{\beta_t^2}{\alpha_t} s(x_t, t) \right) - \dot{\beta}_t \beta_t s(x_t, t).
\end{aligned}
\]

Simplify the coefficients:
\[
\dot{\alpha}_t \left( \frac{\beta_t^2}{\alpha_t} \right) - \dot{\beta}_t \beta_t = \beta_t \left( \frac{\dot{\alpha}_t \beta_t}{\alpha_t} - \dot{\beta}_t \right).
\]

Therefore, the velocity is:
\[
v(x_t, t) = \frac{dx_t}{dt} = \frac{\dot{\alpha}_t}{\alpha_t} x_t + \beta_t \left( \frac{\dot{\alpha}_t \beta_t}{\alpha_t} - \dot{\beta}_t \right) s(x_t, t).
\]
$\square$



\section{Diffusion Generation}
Diffusion models have broad applications in modeling diverse data distributions. Leveraging neural network approximations, these models require substantial amounts of data to enhance their performance. In this section, we first present the pre-training process for various diffusion models discussed in previous sections. Next, we explore methods for diffusion model distillation to enhance inference efficiency. Finally, we outline several approaches for reward-based fine-tuning of these models.
\subsection{Pre-training}
In Section~\ref{sec:diffusion_basic}, we provided detailed introductions to various diffusion models, including the derivation of their training objectives. Here, we present a concise summary of these training objectives in a unified manner for clarity. For more details on each training objective, please refer to the corresponding sections.

\subsubsection{Diffusion Model}

\paragraph{$\epsilon$-prediction Diffusion Model.}
The canonical diffusion model $\epsilon_\theta$ with $\epsilon$-prediction follows the standard objective (omitting coefficients):
\begin{align}
    \mathcal{L}_\text{DM}=\mathbb{E}_{x_0\sim q(x_0), t\in[T], \varepsilon\sim \mathcal{N}(0, \mathbf{I})}[||\epsilon_\theta(x_t, t) - \varepsilon||_2^2]\nonumber
\end{align}
Recall the $x_t$ is a sample with diffusion noise as defined in Eq.~\eqref{eq:x_t}.

Latent diffusion model~\cite{rombach2022high} applies a straightforward method by adding condition variable $c\in\mathcal{C}$ as input argument for conditional generation:
\begin{align}
    \mathcal{L}_\text{LDM}=\mathbb{E}_{x_0\sim q(x_0), c, t\in[T], \varepsilon\sim \mathcal{N}(0, \mathbf{I})}[||\epsilon_\theta(x_t, c, t) - \varepsilon||_2^2]\nonumber
\end{align}
with $x$ being the latent variables after encoding.
The conditional variable is usually also encoded by a text encoder like T5 or CLIP in practice.

\paragraph{$v$-prediction Diffusion Model.}
Alternatively, the diffusion model $v_\theta$ can predict the velocity, known as the $v$-prediction, with the objective:
\begin{align}
        \mathcal{L}_\text{DM-v}=\mathbb{E}_{x_0\sim q(x_0), t\in[T], \varepsilon\sim \mathcal{N}(0, \mathbf{I})}[||v_\theta(x_t, t) - v_t||_2^2]\nonumber
\end{align}
with $v_\theta=\sqrt{1-\bar{\alpha}_t}x_\theta - \sqrt{\bar{\alpha}_t}\epsilon_\theta$, as introduced in Sec.~\ref{sec:v_diff}. This is found to be practically more stable than $\epsilon$-prediction during training in practice~\cite{salimans2022progressive}.

\subsubsection{Consistency Model}
As introduced in Sec.~\ref{sec:consist_train}, the consistency training objective for consistency models $f_\theta$ follows:
\begin{align}
    \mathcal{L}_\text{CT}(\theta)&=\mathbb{E}[\lambda(t_n)d(f_\theta(x_{t_{n+1}}, t_{n+1}),f_{\theta^-}(x_{t_n}, t_{n})]\nonumber
\end{align}
with $d(\cdot, \cdot)$ as a distance metric and $t_n\in[0,T]$ following a continuous time sequence.

\subsubsection{Rectified Flow}
Following previous Sec.~\ref{sec:rectified_flow}, the training objective of rectified flow (RF) model $v_\theta$ is:
\begin{align}
\mathcal{L}_\text{RF}&=\mathbb{E}_{t\in[0,1], x_0\sim q(x_0)}[|| v_\theta(y_t,t) - v_t ||^2]\nonumber\\
v_t&=y_1 - y_0\nonumber \\
y_1&=x_0\nonumber\\
y_0&=\varepsilon, \varepsilon \sim \mathcal{N}(0,\mathbf{I}) \nonumber
\end{align}
In practice, time $t$ is clipped to minimal value like $\sigma_\text{min}=10^{-5}$, and $y_t=t y_1 + (1-(1-\sigma_\text{min})t)y_0, t\in[0,1]$, therefore target velocity is estimated with $v_t=y_1-(1-\sigma_\text{min})y_0$, as in Movie Gen \cite{polyak2024movie}.

Rectified flow is also referred to as flow matching~\cite{lipman2022flow} in literature.

\subsubsection{TrigFlow}
As introduced in Sec.~\ref{sec:trigflow_train}, the TrigFlow model $v_\theta$ has both diffusion training and consistency training objective. The diffusion loss objective follows:
\begin{align}
\mathcal{L}_\text{TF-diffusion}=\mathbb{E}_{x_0, \varepsilon, t}[||\sigma v_\theta(\frac{x_t}{\sigma}, c_\text{noise}(t), c) - v_t||_2^2]\nonumber
\end{align}
with velocity $v_t=\cos(t)\varepsilon - \sin(t)x_0$ for diffusion process.

The consistency training objective has gradients:
\begin{align}
\nabla_\theta\mathcal{L}_\text{TF-CT}=\nabla_\theta \mathbb{E}_{x_t, t}[-\lambda(t)\sin(t)\sigma v^\intercal_\theta(\frac{x_t}{\sigma}, c_\text{noise}(t), c)\frac{d f_{\theta^-}(x_t,t)}{dt}]\nonumber
\end{align}
with continuous-time $t\in[0, \frac{\pi}{2}]$.

\subsection{Distillation}
Vanilla diffusion models typically require a large number of sampling steps through iterative denoising, often ranging from dozens to hundreds of steps in practice, especially for high-dimensional data. Beyond improved model formulations, such as consistency models, rectified flow, or TrigFlow, various techniques exist for distilling these models into versions with fewer sampling steps. This distillation process enhances sampling efficiency while maintaining comparable model performance. In this section, we introduce several such distillation methods.

\subsubsection{Progressive Distillation}
\begin{figure*}[htbp]
    \centering
\includegraphics[width=0.7\textwidth]{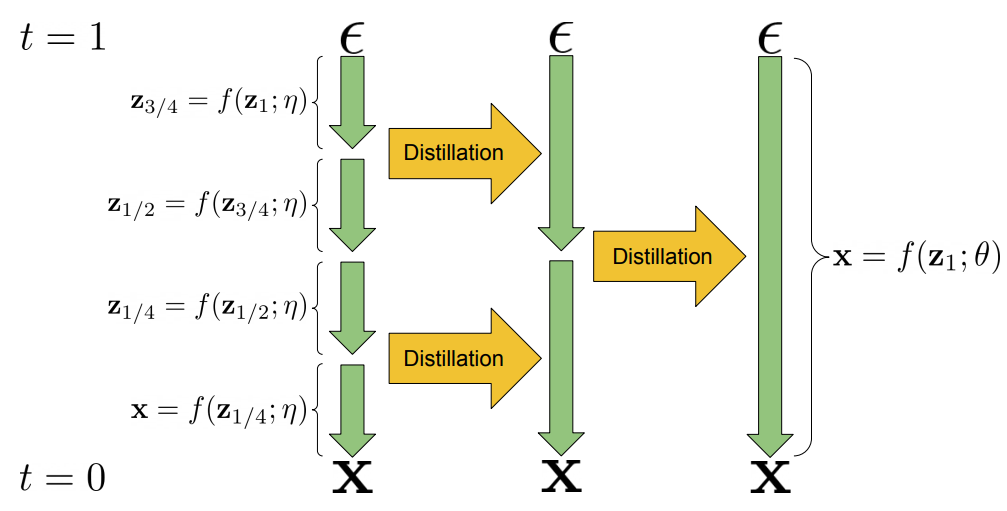}
    \caption{Overview of progressive distillation. Figure is adapted from \cite{salimans2022progressive}.
    }
    \label{fig:progress_distill}
\end{figure*}
As illustrated in Fig.~\ref{fig:progress_distill},
\textbf{Progressive Distillation} (PD) \cite{salimans2022progressive} iteratively halves the sampling steps of a teacher model and distill it into a student model. For example, for a teacher model trained on time sequence $\{t_n\}, n\in\{0, 0.5, 1, \dots,N-1, N-0.5, N\}$, the student will follow time sequence $\{t_n\}, n\in\{0, 1, \dots, N\}$.
Following Eq.~\eqref{eq:x_t_1_3_general}, we define a single DDIM step as:
\begin{align}
    x_{t_{n-1}}(x_{t_n}, \epsilon_\theta, \varepsilon)&=\sqrt{\bar{\alpha}_{t_{n-1}}}\big(\frac{x_{t_n}-\sqrt{1-\bar{\alpha}_{t_n}}\epsilon_\theta}{\sqrt{\bar{\alpha}_{t_n}}}\big) + \sqrt{1-\bar{\alpha}_{t_{n-1}}-\sigma_{{t_n}}^2}\epsilon_\theta + \sigma_{t_n} \varepsilon \label{eq:ddim}\\
   &\triangleq\text{DDIM}(x_{t_n}, \epsilon_\theta)\nonumber
\end{align}
To distill from a teacher model $\epsilon_{\theta'}$ with $2N$ steps, one iteration of PD will distill it into a student model $\epsilon_\theta$ with $N$ steps. Specifically,
the loss function is
\begin{align*}
    \mathcal{L}_\text{PD}(\theta)&=\mathbb{E}_{i\in[N]}||\tilde{x}_{t_{i-1}} - \hat{x}_{t_{i-1}}||^2 \\
    \text{with,} \\
    \tilde{x}_{t_{i-0.5}}&=\text{DDIM}(x_{t_i}, \epsilon_{\theta'})\\
    \tilde{x}_{t_{i-1}}&=\text{DDIM}(x_{t_{i-0.5}}, \epsilon_{\theta'})\\
    \hat{x}_{t_{i-1}}&=\text{DDIM}(x_{t_i}, \epsilon_{\theta})\\
\end{align*}
The process matches the student’s one-step denoising prediction with the teacher’s two-step denoising prediction. In each iteration, the student model from the current step serves as the teacher for the next distillation iteration. After $M$ iterations of the PD process, the student achieves a sampling process with $\lceil \frac{N}{2^M} \rceil$ steps.

\textbf{Guided-PD}~\cite{meng2023distillation} extends the progressive distillation (PD) process by incorporating classifier-free guidance (CFG) from the teacher, enabling conditional generation. Specifically, the teacher leverages CFG during the distillation process. Recall Eq.~\eqref{eq:cfg} for the CFG-augmented prediction with guidance weight $w$:
\begin{align*}
    \epsilon_\theta(x_{t_{n+1}}, c, t_{n+1}, t_n,  w) = \epsilon_\theta(x_{t_{n+1}}, c, t_{n+1}, t_n) + w(\epsilon_\theta(x_{t_{n+1}}, c, t_{n+1}, t_n)- \epsilon_\theta(x_{t_{n+1}}, \varnothing, t_{n+1}, t_n))
\end{align*}
Replacing it into Eq.~\eqref{eq:ddim} gives one-step DDIM inference with CFG: $\text{DDIM}^\text{CFG}(x_{t_n}, \epsilon_\theta, c, w)$ with conditional variable $c\in\mathcal{C}$. Practical guidance weight applies a constant value or a value uniformly sampled from a pre-specified range. The rest of Guided-PD follows standard PD.

\subsubsection{Score Distillation Sampling}
Following the previously introduced score matching objective in Sec.~\ref{sec:diff_train}, there is a series of work for model distillation via score matching. 

DreamFusion~\cite{poole2022dreamfusion} introduces Score Distillation Sampling (SDS) for maximum likelihood estimation in 3D generation.
The objective for SDS is to minimize the KL-divergence of parameterized noised sample distribution $p_t^\theta(x_t|c)$ at timestep $t$ along the forward diffusion process and the real data distribution $q_t(x_t|c)$ for $\forall t\in[0.02T, 0.98T]$:
\begin{align}
    \mathcal{L}_\text{SDS}=\mathbb{E}_{t,c, \varepsilon\sim\mathcal{N}(0, \mathbf{I})}[D_\text{KL}(p^\theta_t(x_t|c)||q_t(x_t|c))]\nonumber
\end{align}
$T$ is the time horizon of the diffusion process (\emph{e.g.}, 1 for continuous timesteps or 1000 for discrete timesteps). The timestep is clipped because the score estimation can be unstable at the time boundaries, especially when $t\rightarrow 0$.

The gradient of $\mathcal{L}_\text{SDS}$ is, 
\begin{align*}
   \nabla_\theta \mathcal{L}_\text{SDS}&=\mathbb{E}_{t, c,\varepsilon\sim\mathcal{N}(0, \mathbf{I})}\big[w(t)\big(\nabla_{x_t}\log p_t^\theta(x_t|c) - \nabla_{x_t}\log q_t(x_t|c)\big)\nabla_\theta G_\theta(\varepsilon, c)\big]\\
   &= \mathbb{E}_{t, c,\varepsilon\sim\mathcal{N}(0, \mathbf{I})}[w(t)(-\frac{\epsilon_\theta}{\sqrt{1-\bar{\alpha}_t}} - (-\frac{\varepsilon}{\sqrt{1-\bar{\alpha}_t}}) )\nabla_\theta G_\theta(\varepsilon, c)]
\end{align*}
by applying the score estimation Eq.~\eqref{eq:score_epsilon} in DDPM schedule.

\subsubsection{Variational Score Distillation}
ProlificDreamer~\cite{wang2024prolificdreamer} proposes Variational Score Distillation (VSD) for 3D generation with diffusion models, as a generalized version of SDS. As pointed out in the paper, SDS often suffers from oversaturation, over-smoothing, and low-diversity issues.

Considering a parameterized generator $G_\theta(c)$, with its parameters sample from distribution $\mu(\theta)$, suppose the ground truth real data sample from $q(x_0|c)$ with $c$ being conditions, the variational score distillation (VSD) solves the following variational inference problem,
\begin{align}
    \mu^*=\min_\mu D_\text{KL}(p^\mu_0(x_0|c)||q_0(x_0|c))\nonumber
\end{align}
For a complex data distribution $q_0$, this can be hard to optimized. However, it can be easier to optimize the problem for the distribution $q_t(x_t|c)$ of diffused sample $x_t$ with larger $t$ via a diffusion process. Thus, VSD construct an alternative objective for optimization with KL-divergence of denoising distributions over all timesteps,
\begin{align}
    \mu^*=\min_\mu \mathbb{E}_{t,c}[\frac{1}{\sqrt{\bar{\alpha}_t}}w(t)D_\text{KL}(p^\mu_t(x_t|c)||q_t(x_t|c))]
    \label{eq:vsd_obj}
\end{align}
by leveraging the diffusion probability $q(x_t|x_0)$ as Eq.~\eqref{eq:xt_x0}. It can be proved that Eq.~\eqref{eq:vsd_obj} yields the same optimal $\mu^*$ as previous objective, with the following theorem.
\begin{theorem}[Global optimum of VSD] For $\forall t>0$,
\begin{align*}
    D_\text{KL}(p^\mu_t(x_t|c)||q_t(x_t|c))=0\Leftrightarrow p^\mu_0(x_0|c)||q_0(x_0|c)
\end{align*}
\end{theorem}
The above alternative objective actually traces back to the denoising score matching~\cite{vincent2011connection} and has been discussed in later work~\cite{song2019generative}. 

The gradient of the VSD objective is:
\begin{align}
    \nabla_\theta \mathcal{L}_\text{VSD}=\mathbb{E}_{t, c,\varepsilon\sim\mathcal{N}(0, \mathbf{I})}[w(t)(\nabla_{x_t}\log p_t^\mu(x_t|c) - \nabla_{x_t}\log q_t(x_t|c))\nabla_\theta G_\theta(\varepsilon, c)]\nonumber
\end{align}
with $\nabla_\theta x_t = \sqrt{\bar{\alpha}_t}\nabla_\theta \hat{x}_0=\sqrt{\bar{\alpha}_t}\nabla_\theta G_\theta$, $\sqrt{\bar{\alpha}_t}$ absorbed in $w(t)$. $G_\theta(\varepsilon, c)$ is a conditional variational generator depending on both the random noise $\varepsilon\sim \mathcal{N}(0, \mathbf{I})$ and conditional variable $c$. $\nabla_{x_t}\log q_t(x_t|c)$ is the real score function on noisy generated images $x_t$, which is estimated with a pre-trained diffusion model on real samples.
$\nabla_{x_t}\log p_t^\mu(x_t|c)$ is the fake score function on noisy generated images, which is estimated with another diffusion model trained via the standard diffusion objective on generated sample distribution by $G_\theta$ as:
\begin{align}
    \mathcal{L}_\text{diffusion}(\phi)=\mathbb{E}_{x_0, t,\varepsilon}||\varepsilon - \epsilon_\phi(F(G_\theta(c), t, \varepsilon))||^2
    \label{eq:vsd_diff}
\end{align}
with $F(\cdot)$ as the forward diffusion process.

It needs to notice that the \textit{global optimum of VSD} indicates the condition of minimal difference of scores for $\forall t>0$ is sufficient for achieving the minimal different of scores at $t=0$, but it does not indicate that each score can be estimated in this way individually.

More concretely, the true score can not be approximated with the expectation of diffused score estimation over diffusion timesteps and perturbation noise:
\begin{align}
    s(x_0) \neq \mathbb{E}_{t, x_t, \epsilon}[s(x_t, t)]
    \label{eq:dmd_score}
\end{align}

\subsubsection{Distribution Matching Distillation}
\begin{figure*}[htbp]
    \centering
\includegraphics[width=0.99\textwidth]{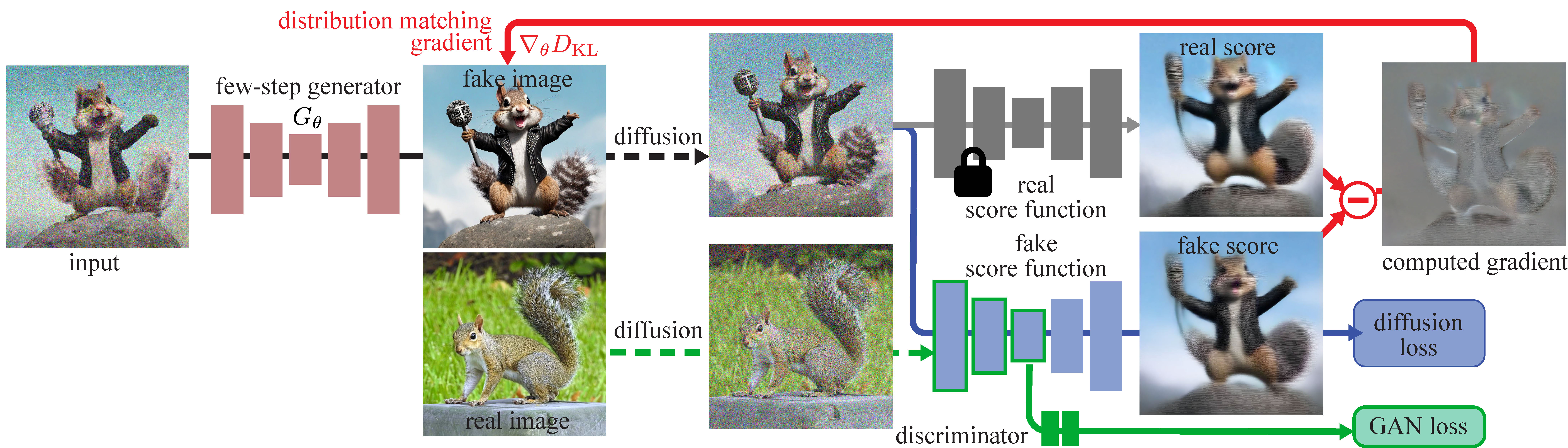}
    \caption{Overview of distribution matching distillation. Figure is adapted from \cite{yin2024improved}.
    }
    \label{fig:dmd}
\end{figure*}
As illustrated in Fig.~\ref{fig:dmd}, the distribution matching distillation (DMD)~\cite{yin2024one, yin2024improved} is essentially VSD with additional generative adversarial network (GAN) loss for discriminating between the generated samples and training samples:
\begin{align}
    \mathcal{L}_\text{DMD}= \mathcal{L}_\text{VSD}(\theta)+\lambda_1 \mathcal{L}_\text{diffusion}(\phi)+\lambda_2 \mathcal{L}_\text{adv}(\theta, \psi)\nonumber
\end{align}
DMD is employed to distill knowledge from a teacher diffusion model into a few-step student diffusion model, which, in this case, serves as the parameterized generator $G_\theta$. Unlike VSD, which directly optimizes the $\theta$-parameterized 3D generation~\cite{wang2024prolificdreamer}, the DMD loss optimizes an individual diffusion network for $G_\theta$.

\paragraph{Variational Score Distillation Loss.}
Similar as in \cite{wang2024prolificdreamer}, the variational score distillation (VSD) loss is designed to minimize the Kullback-Leibler (KL) of real (teacher) sample distribution $p_\text{real}$ and fake (student) sample distribution $p_\text{fake}$, specifically:
\begin{align*}
    \mathcal{L}_\text{VSD}&\coloneqq D_\text{KL}(p_\text{fake}||p_\text{real})=\mathbb{E}_{x\sim p_\text{fake}}[\log \frac{p_\text{fake}(x)}{p_\text{real}(x)}]\\
    &=\mathbb{E}_{\varepsilon\sim \mathcal{N}(0, \mathbf{I}), x=G_\theta(\varepsilon)}[\log \frac{p_\text{fake}(x)}{p_\text{real}(x)}]\\
\end{align*}
and the derivative of the objective is,
\begin{align}
    \nabla_\theta D_\text{KL}=\mathbb{E}_{\varepsilon\sim \mathcal{N}(0, \mathbf{I}), x=G_\theta(\varepsilon)}[-(s_\text{real}(x)-s_\text{fake}(x))\nabla_\theta G_\theta(\varepsilon)]
    \label{eq:dmd_vsd}
\end{align}
with score functions $s_\text{real}(x)=\nabla_x \log p_\text{real}(x)$ and $s_\text{fake}(x)=\nabla_x \log p_\text{fake}(x)$ for two distributions. For practical estimation of the scores, the samples need to be diffused with noise to be $x_t$ via forward diffusion. This gradient is essentially the same as the gradient of VSD objective. 
\textit{The above scores are estimated with the diffusion model on their diffused distributions, as a sufficient condition for matching the corresponding scores on original real or fake distributions, according to the global optimum of VSD.}

Specifically, according to Eq.~\eqref{eq:score_x_t}, given diffused distribution $q(x_t|x_0)\sim\mathcal{N}(x_t;\sqrt{\bar{\alpha}_t}x_0, (1-\bar{\alpha}_t)\mathbf{I})$, the score $s_\theta=\nabla_{x_t}\log q(x_t)$ can be estimated with the diffusion model by:
\begin{align}
    s_\theta(x_t, t) = -\frac{x_t-\sqrt{\bar{\alpha}_t}x_\theta}{1-\bar{\alpha}_t}
    \label{eq:score_dmd}
\end{align}
where $x_\theta=\frac{1}{\sqrt{\bar{\alpha}_t}}x_t-\frac{\sqrt{1-\bar{\alpha}_t}}{\sqrt{\bar{\alpha}_t}}\epsilon_\theta$ is the reparameterization from $\epsilon$-prediction to $x_0$-prediction, using Eq.~\eqref{eq:xt_x0}. There is also a shortcut to directly using $s_\theta(x_t, t)=-\frac{\epsilon_\theta(x_t, t)}{\sqrt{1-\bar{\alpha}_t}}$ as Eq.~\eqref{eq:score_epsilon}, which yields the same results.

The practical operations for deriving the gradient update with VSD loss are:
\begin{enumerate}
\item Given data sample $x_0$ generated with $G_\theta$, add noise $\varepsilon$ to $x_{\tau}(\varepsilon)$ through forward diffusion as Eq.~\eqref{eq:xt_1_x0}.
\item  The student generator predicts denoised sample $x^f=G_\theta(x_{\tau}(\varepsilon))$, then forward diffuses it to $x^f_{t}(\varepsilon')$ with noise $\varepsilon'$ with the same forward diffusion equation. Two different timestep sequences are applied. $\tau\in\{249, 499, 749, 999\}$ follows $4$-step student sampling timesteps, while $t\in\{1,2,\dots,999\}$ follows teacher sampling timesteps. 
\item 
Now calculate the real and fake score for $x^f_{t}(\varepsilon')$ using aforementioned score equation Eq.~\eqref{eq:score_dmd}.

The real score $s^r(x^f_t)$ is computed using a frozen teacher model, while the fake score $s^f_\phi(x^f_t)$ is obtained from a copy of the teacher model with adaptive learning. The fake score model is updated through the diffusion loss during the distillation process to align with the distribution of the student-generated samples.

\item Take stochastic gradient descent to update the generator. Following Eq.~\eqref{eq:dmd_vsd}, the gradient of distillation matching distillation loss has format:
\begin{align*}
    \nabla_\theta \mathcal{L}_\text{VSD}(\theta)&=\mathbb{E}_{t, x_0, \varepsilon}\big[w(t)\big(s_\phi^f(x^f_{t}, t) - s^r(x^f_{t}, t)\big)\big]\nabla_\theta G_\theta(x_{\tau}(\varepsilon))\\
    &=\nabla_\theta \mathbb{E}_{t, x_0, \varepsilon}\big[\frac{w(t)}{2}|| x^f - [x^f - \big(s_\phi^f(x^f_{t}, t) - s^r(x^f_{t}, t)\big)]\text{.detach()}||^2]\\
    &=\nabla_\theta \mathbb{E}_{t, x_0, \varepsilon}\big[\frac{w(t)}{2}\frac{\sqrt
    {\bar{\alpha}_t}}{1-\bar{\alpha}_t}||x^f - [x^f - \big(x_\phi^f(x_t^f, t) - x^r(x_t^f, t)]\text{.detach()}\big) ||^2\big]
\end{align*}
where $x_f=G_\theta(x_{\tau}(\varepsilon))$ and $w(t)$ is a time-dependent coefficient. The last equation is derived by score function Eq.~\eqref{eq:score_dmd}, with $x_\phi^f(x_t^f, t)$ using the fake score model and $x^r(x_t^f, t)$ using the teacher model.  The symbol ``.detach()'' indicates stopping the gradient in practice. In practice the weight coefficient is chosen as $w(t)=\frac{2(1-\bar{\alpha}_t)}{\sqrt{\bar{\alpha}_t}}\frac{c}{||x_f -  x^r(x_t^f, t)||_1}$ with $c$ as a constant of spatial and channel dimensions for better performances.
\end{enumerate}

\paragraph{Diffusion Loss.}
Same as Eq.~\eqref{eq:vsd_diff}, the diffusion loss for fake score network $s^f_\phi$ follows standard diffusion model training for $\epsilon$-prediction, $x$-prediction or $v$-prediction. Here we take $\epsilon$-prediction as an example. The diffusion loss has form:
\begin{align}
    \mathcal{L}_\text{diffusion}(\phi)=\mathbb{E}_{x_0, t,\varepsilon}||\varepsilon - \epsilon^f_\phi(F(x^f, t, \varepsilon))||^2\nonumber
\end{align}
$x^f_{t}(\varepsilon')=F(x^f, t, \varepsilon')$ with $F$ as forward diffusion process.
From Eq.~\eqref{eq:score_epsilon}, we know $\epsilon^f_\phi$ and score $s^f_\phi$ can be mutually transformed by multiplying a simple coefficient.

\paragraph{Adversarial Loss.}
An additional GAN-type adversarial loss is introduced, leveraging a learned discriminator $D_\psi$.
The adversarial loss has form:
\begin{align}
    \min_\theta \max_\psi \mathcal{L}_\text{adv}(\theta, \psi)=\mathbb{E}_{x'_0, t'\in[T], \varepsilon}[\log D_\psi(F(x'_0, t', \varepsilon)] + \mathbb{E}_{x_0, t\in[T], \varepsilon'}[-\log D_\psi(x^f_t(\varepsilon'))]\nonumber
\end{align}
The forward diffusion function $F(x'_0, t', \varepsilon)$ generates a noisy sample from a clean sample $x'_0$ in the dataset, where $\varepsilon \sim \mathcal{N}(0, \mathbf{I})$ represents forward diffusion noise. The student-generated sample with noise is expressed as $x^f_t(\varepsilon') = F(x^f, t, \varepsilon') = F(G_\theta(x_\tau(\varepsilon)), t, \varepsilon')$. Note that the notations $x'_0$ and $x_0$, $t'$ and $t$, as well as $\varepsilon$ and $\varepsilon'$, indicate that each pair can represent distinct values. The discriminator $D_\psi$ shares the first half of the fake score network $s_\phi$ as its backbone. Alternatively, it could share the backbone with the real score network (frozen), but this approach would result in only the discriminator head being updated, which may be insufficient for learning an accurate discriminator. Here, $t, t' \in \{1, 2, \dots, 999\}$.

\paragraph{Inference.}
Following the inference method of the consistency model, the process iteratively alternates between denoising and adding noise through forward diffusion, following a reverse sequence of timesteps $\{999, 749, 499, 249\}$ for 4-step sampling:
\begin{align*}
    \hat{x}_{0}&=G_\theta(x_{t_n}, t_n) \quad\quad\quad\quad\quad\quad\quad\quad\quad\text{(denoising)}\\
    x_{t_{n-1}}&=\sqrt{\bar{\alpha}_{t_{n-1}}}\hat{x}_0+\sqrt{1-\bar{\alpha}_{t_{n-1}}}\varepsilon \quad\quad\quad \text{(forward diffusion)}
\end{align*}

\subsubsection{Adversarial Diffusion Distillation}
\begin{figure*}[htbp]
    \centering
\includegraphics[width=0.6\textwidth]{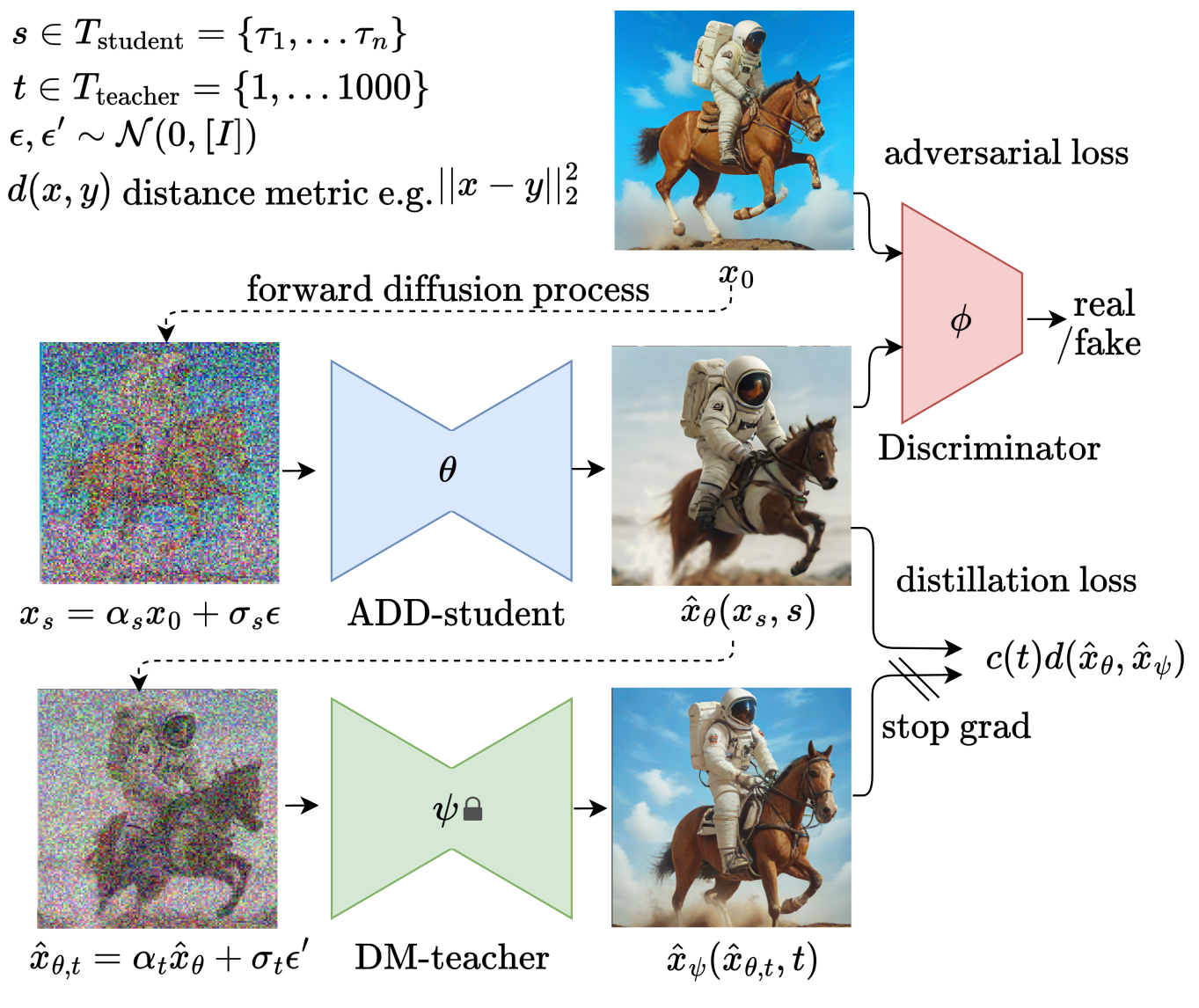}
    \caption{Overview of adversarial diffusion distillation. Figure is adapted from \cite{sauer2023adversarial}.
    }
    \label{fig:add}
\end{figure*}
As displayed in Fig.~\ref{fig:add}, adversarial diffusion distillation (ADD)~\cite{sauer2023adversarial} combines adversarial loss and distillation loss:
\begin{align}
    \mathcal{L}_\text{ADD}=\mathcal{L}_\text{adv} + \lambda \mathcal{L}_\text{distill}\nonumber
\end{align}

The adversarial loss follows standard GAN method, with a discriminator $D_\phi$ to distinguish between the student generated samples and real samples $x_0$:
\begin{align}
    \min_\theta \max_\phi L_\text{adv}(\theta, \phi)=\mathbb{E}_{\varepsilon, s\in\{\tau_i\}_{i=1}^n, x_0}[\log D_\phi(\hat{x}_\theta(x_s, s)) + \log (1-D_\phi(x_0))]\nonumber
\end{align}

The distillation loss is to matching student and teacher prediction across timesteps:
\begin{align}
    \mathcal{L}_\text{distill}=\mathbb{E}_{t\in[T], s\in\{\tau_i\}_{i=1}^n}[c(t)d(\hat{x}_\theta(x_s, s), \hat{x}_{\psi}(x_t, t))]\nonumber
\end{align}
where the student follows a sub-sequence of the teacher timesteps $\{\tau_i\}_{i=1}^n\subseteq [T]$, and $d(\cdot, \cdot)$ is a distance metric like MSE. $\hat{x}_\theta(x_s, s)$ and $\hat{x}_{\psi}(x_t, t)$ are predictions from student network $\epsilon_\theta$ and teacher network $\epsilon_{\psi}$ respectively, following Eq.~\eqref{eq:tweedie}.

\subsubsection{Consistency-based Distillation}
\label{subsec:cd}
The consistency model introduced in Sec.~\ref{sec:consistency_model} can be applied for distillation from pre-trained diffusion model to reduce sampling steps. 
\paragraph{Consistency Distillation.}
Consistency model \cite{song2023consistency} enforces the consistency loss as the distillation method from a pre-trained teacher model $\epsilon_{\theta'}$, with a discrete sub-sampled time schedule $t_1=\epsilon< t_2<\dots < t_N=T$:
\begin{align}
    \mathcal{L}_\text{CD}(\theta)&=\mathbb{E}[\lambda(t_n)d(f_\theta(x_{t_{n+1}},c, t_{n+1}),f_{\theta^-}(\hat{x}_{t_n},c, t_{n})]\label{eq:cd_loss}\\
    \hat{x}_{t_n}&=\text{Denoise}(x_{t_{n+1}}, c, t_{n+1}, t_n; \theta')
    \nonumber
\end{align}
where $\lambda(t_n)$ is a time dependent coefficient usually set as constant in practice, and $d(\cdot, \cdot)$ is a distance metric like MSE or Huber loss. One example of the one-step denoising function $\text{Denoise}(\cdot)$ is Eq.~\eqref{eq:ode_ddim} for DDIM sampling with a pre-trained noise prediction diffusion model.

Latent Consistency Model (LCM)~\cite{luo2023latent} introduces a way to reparameterize the $f$-prediction in consistency model with $\epsilon$-prediction. Specifically, with a pre-trained teacher diffusion model $\epsilon_{\theta'}$, the student consistency model $f_\theta$ can be parameterized according to Eq.~\eqref{eq:parameterize_cm_dm}:
\begin{align*}
    f_\theta(x_t,c,t) &= c_{\text{skip}} x_t + c_{\text{out}}x_\theta(x_t, c, t)\\
    x_\theta&=\frac{x_t-\sqrt{1-\bar{\alpha}_t}\epsilon_\theta(x_t,c,t)}{\sqrt{\bar{\alpha}_t}}
\end{align*}
This allows student network $\epsilon_\theta$ to be initiated with the teacher model $\epsilon_{\theta'}$ for leveraging the network compressed information during pre-training. If the student predicts other variables instead of the noise, the above reparameterization will take another form (see transformation between parameterization in Sec.~\ref{sec:parameterization}). $c$ is the condition variable in teacher network for conditional generation. To further enhance the conditioning capability, the classifier-free guidance (CFG) is applied:
\begin{align}
    \epsilon_\theta(x_t, c, t)=(1+w)\epsilon_\theta(x_t, c, t) - w \epsilon_\theta(x_t, \varnothing, t)\nonumber
\end{align}

The distilled student network also supports conditional generation with variable $c$ but without CFG at inference. 

This technique is applied in LCM~\cite{luo2023latent}, Animate LCM~\cite{wang2024animatelcm}, etc. Due to distilling from the teacher diffusion model, LCM methods drop the original noise schedule of CM but follow the noise schedule of teacher DM. Also, the student CM can take any subsequence $\{t_n\}_{n=1}^N$ of original timesteps $[T]$ during distillation and inference, which is natively supported by CM to achieve acceleration with reduced sampling steps. The inference pseudo-code is shown in Alg.~\ref{alg:lcm_inference}.

\begin{algorithm}[H]
\caption{LCM Inference}
\begin{algorithmic}[1]
\State $x_{t_N}\sim\mathcal{N}(0, \mathbf{I}), \{t_n\}_{n=1}^N\subseteq[T]$
\For{$n \in [N]$}
    \State predict noise $\epsilon_\theta(x_{t_n}, c,t_n)$
    \State $\hat{x}_0 \gets f_\theta(x_{t_n}, c, t_n)$ by consistency prediction Eq.~\eqref{eq:parameterize_cm_dm}
    \State $x_{t_{n-1}} \gets \sqrt{\bar{\alpha}_{t_{n-1}}}\hat{x}_0 + \sqrt{1-\bar{\alpha}_{t_{n-1}}} \varepsilon,  \varepsilon \sim \mathcal{N}(0,\mathbf{I})$
\EndFor
\end{algorithmic}
\label{alg:lcm_inference}
\end{algorithm}

\paragraph{Generalized CD with multi-step denoising.}
The one-step CD loss can be generalized to $m$-step CD loss in practice, which changes Eq.~\eqref{eq:cd_loss} to be:
\begin{align}
    \mathcal{L}_\text{CD}(\theta)&=\mathbb{E}_{x_0\sim q(x_0), t_n}[\lambda(t_n)d(f_\theta(x_{t_{n+m}}, c,t_{n+m}),f_{\theta^-}(\hat{x}_{t_n}, c,t_{n})]\label{eq:general_cd_loss}\\
    \hat{x}_{t_n}&=\text{Denoise}^m(x_{t_{n+m}},c, t_{n+m}, t_n; \theta')\nonumber
\end{align}
 with $\text{Denoise}^m(\cdot)$ indicating the $m$-step denoising function as defined by Eq.~\eqref{eq:multi_step_denoise}, which iteratively predicts the sequence $(\hat{x}_{t_{n+m-1}}, \dots, \hat{x}_{t_{n}}|x_{t_{n+m}})$.

\paragraph{Consistency Training for Distillation.}
We may also wonder if consistency training (CT), as another method for training consistency models, can help with the distillation process. The CT loss has the form:
\begin{align}
    \mathcal{L}_\text{CT}(\theta)&=\mathbb{E}[\lambda(t_n)d(f_\theta(x_{t_{n+1}}, t_{n+1}),f_{\theta^-}(x_{t_n}, t_{n})]\nonumber
\end{align}
with $x_{t_{n+1}} = \sqrt{\bar{\alpha}_{t_{n+1}}}x_0+\sqrt{1-\bar{\alpha}_{t_{n+1}}}\varepsilon, x_{t_{n}} = \sqrt{\bar{\alpha}_{t_{n}}}x_0+\sqrt{1-\bar{\alpha}_{t_{n}}}\varepsilon, \varepsilon\sim\mathcal{N}(0, \mathbf{I})$, as forward diffusion process.

If the timestep schedule follows LCM with a constant interval $|t_{n+1}-t_n|= C$, one problem of above CT method for distillation is that it is not theoretically guaranteed to achieve a converging loss. 
From Theorem~\ref{theorem:ct} we know, if $|t_{n+1}-t_n|\le C$, 
then
\begin{align*}
&\mathcal{L}^N_\text{CD}=\mathcal{L}^N_\text{CT} + o(C)\\
&\mathcal{L}^N_\text{CT}\geq O(C) \text{ if } \inf \mathcal{L}^N_\text{CD}>0
\end{align*}
which yields a constant error $O(C)$ for $\mathcal{L}^N_\text{CT}$. Additionally, $o(C)$ is not asymptotically converging to 0 as $C$ is not small (the Taylor expansion may not hold). This makes both CT and CD loss at least a constant error in theory. For above CT to converge, it requires a timestep schedule with asymptotically decreasing interval $|t_{n+1}-t_n|\le o(C)$ as training iteration increases.
LCM applies CD which directly minimizes $\mathcal{L}^N_\text{CD}$, therefore a constant timestep interval does not affect its convergence. 

People may ask, instead of matching $f_\theta(x_{t_{n+1}}, t_{n+1})$ and $f_{\theta^-}(x_{t_n}, t_{n})$, which accumulates error as $t_n$ becomes large, why not directly matching $f_{\theta}(x_{t_n}, t_{n})$ with $x_0$. It follows the loss:
\begin{align}
    \mathcal{L}(\theta)&=\mathbb{E}[\lambda(t_n)d(f_\theta(x_{t_n}, t_n),x_0)], n\in[N]\nonumber
\end{align}
This actually corresponds to the diffusion loss for $x$-prediction, which can be readily proven. However, directly minimizing $d(f_\theta(x_{t_n}, t_n), x_0)$ does not require teacher guidance, and thus does not constitute distillation from any teacher model. Without leveraging the guidance of the teacher along the diffusion path, training to predict $f_\theta(x_{t_n}) \to x_0$ becomes a challenging task, as the noise $\epsilon_\theta$ to predict can vary significantly across different $x_{t_n}$. The teacher's denoising function provides intermediate targets for the student to learn along the diffusion path, thereby alleviating the difficulty of learning (as a hypothetical statement).

To enable distillation from a pretrained teacher model with parameters $\theta'$, the target $x_0$ in the above loss can be replaced with the teacher's predicted $\hat{x}_0 = \text{Denoise}^n(x_{t_n})$, which requires an $n$-step denoising process. While this approach introduces potentially high computational costs, it provides a regression loss on teacher-generated samples, analogous to the regression loss used in distribution matching distillation~\cite{yin2024one}.

\paragraph{Student-Teacher Parameterization.}  
Student-teacher parameterization can be categorized into two types: \textbf{homogeneous} and \textbf{heterogeneous}.

For homogeneous student-teacher parameterization, the student and teacher networks follow the same variable prediction, \emph{e.g.}, both utilizing $\epsilon$-prediction as in the LCM case described in Sec.~\ref{subsec:cd}.

For heterogeneous student-teacher parameterization, the student and teacher networks adopt different variable predictions, \emph{e.g.}, $\epsilon$-prediction for the teacher and $v$-prediction for the student. This setup necessitates an additional transformation between parameterizations, as discussed in Sec.~\ref{sec:parameterization}.

For both types of parameterization, the student model can be initialized from the teacher model at the start of the distillation process, provided the network architectures are identical. However, homogeneous parameterization is expected to preserve more coherent weight information theoretically.

\subsection{Reward-based Fine-tuning}
Similar as reinforcement learning from human feedback (RLHF) in large language models, reward models can be used for fine-tuning pre-trained diffusion models. 
Based on whether the reward model's gradient is utilized for optimization, reward fine-tuning methods can be categorized into two types: (1). direct reward gradient and (2). gradient-free reward optimization.

\subsubsection{Direct Reward Gradient}

We first introduce three methods requiring a differentiable reward model for gradient-based optimization, \emph{i.e.}, $\nabla_x R(x)$, which includes ReFL, DRaFT and Q-score matching.

\paragraph{Reward Feedback Learning (ReFL)~\cite{xu2024imagereward}.}
ReFL applies reward gradient for diffusion model optimization. It backpropagates the reward gradient through one-step predicted $x_\theta=\frac{1}{\sqrt{\bar{\alpha}_t}}x_t-\frac{\sqrt{1-\bar{\alpha}_t}}{\sqrt{\bar{\alpha}_t}}\epsilon_\theta$ as Eq.~\eqref{eq:tweedie}, similar as diffusion posterior sampling~\cite{chung2022diffusion}. The reward-based diffusion fine-tuning objective is:
\begin{align}
    \nabla_\theta \mathcal{L}_\text{ReFL}&= -\mathbb{E}_{\varepsilon\sim \mathcal{N}(0, \mathbf{I})}[\nabla_\theta R_\phi(x_\theta)]\nonumber\\
     &=-\mathbb{E}_{\varepsilon\sim \mathcal{N}(0, \mathbf{I})}[\nabla_{x_\theta} R_\phi(x_\theta)\nabla_{\theta}x_\theta]\nonumber
\end{align}
with $R_\phi$ as a differentiable reward model and $\epsilon_\theta$ as the noise-prediction diffusion model.

\paragraph{Direct Reward Fine-Tuning (DRaFT)~\cite{clark2023directly}.}
DRaFT requires the reward gradient to optimize the diffusion model, which optimize the following:
\begin{align}
    \nabla_\theta \mathcal{L}_\text{DRaFT}&= -\mathbb{E}_{\varepsilon\sim \mathcal{N}(0, \mathbf{I})}[\nabla_\theta R_\phi(G_\theta(\varepsilon))]\nonumber\\
    &=-\mathbb{E}_{\varepsilon\sim \mathcal{N}(0, \mathbf{I})}[\nabla_x R_\phi(x)\nabla_\theta G_\theta(\varepsilon)]\nonumber
\end{align}
where $x=G_\theta(\varepsilon)$ requires iterative sampling through the denoising function from noise $\varepsilon$.
In practice, DRaFT usually requires truncation on gradient backpropagation along the diffusion chain, and find that truncating DRaFT to a single backwards step substantially improves sample efficiency. DRaFT-$K$ truncates the reward gradient backpropagation in diffusion process to latest $K$ steps as a trade-off between efficiency and performance.

\paragraph{Q-score Matching~\cite{psenka2023learning}.}
The Q-score matching (QSM) proposes to directly match the score prediction from the diffusion model with the sample gradient from the value function (either reward or a Q-value in RL setting),
\begin{align}
    \mathcal{L}_\text{QSM}(\theta)=\mathbb{E}_{\varepsilon\sim\mathcal{N}(0, \mathbf{I}), t, x_t}||s_\theta(x_t, t) - \nabla_{x_t} R(x_t)||^2\nonumber
\end{align}
with $s_\theta(x_t, t)=-\frac{x_t-\sqrt{\bar{\alpha}_t}\hat{x}_{\theta}}{1-\bar{\alpha}_t}$.
During usage, it requires to estimate reward values for $x_t$ across all timesteps, $\forall t \in [T]$. This task can be ill-posed for the reward model, as it is typically designed to operate within the clean sample space, free from noise, \emph{i.e.}, at $t=0$.

\subsubsection{Gradient-free Reward Optimization}
Different from previous methods requiring reward gradients, we introduce other methods as ``zero-th order'' optimization without directly leveraging the reward gradients.

\paragraph{Denoising Diffusion Policy Optimization (DDPO)~\cite{black2023training}.}
Also inspired from REINFORCE algorithm, DDPO follows:
\begin{align}
     \nabla_\theta \mathcal{L}_\text{DDPO}&=-\mathbb{E}_{x_0\sim p_\theta(x_0)}[\nabla_\theta \log p_\theta (x_0) R(x_0)]\nonumber\\
     &=-\mathbb{E}_{\tau_\sim p_\theta}[\sum_{t=1}^T \nabla_\theta \log p_\theta (x_{t-1}|x_t) R(x_0)]\label{eq:ddpo_sf}\\
     &(\text{with } p_\theta(x_0)=\int_{x_1}\dots\int_{x_T}\Pi_{t=1}^T p_\theta(x_{t-1}|x_t) d x_1 d x_2 \dots d x_{T})\nonumber
\end{align}
where $\tau=\{x_T, x_{T-1}, \dots, x_0\}, x_T\sim\mathcal{N}(0, \mathbf{I})$.
Notice that this requires to additionally marginalize out all intermediate variables $\tau_{T:1}=\{x_T, x_{T-1}, \dots, x_1\}$ generated along the diffusion chain, by the integral in expectation.
In DDPM, the $p_\theta (x_{t-1}|x_t)$ is isotropic Gaussian following Eq.~\eqref{eq:q_x_t_1}:
\begin{align}
    q(x_{t-1}|x_t, x_0)
    &\propto \mathcal{N}\big(x_{t-1}; \frac{\sqrt{\alpha_t}(1-\bar{\alpha}_{t-1})x_t+\sqrt{\bar{\alpha}_{t-1}}(1-\alpha_t)x_0}{1-\bar{\alpha}_t}, \frac{(1-\alpha_t)(1-\bar{\alpha}_{t-1})}{1-\bar{\alpha}_t}\mathbf{I}\big)\nonumber
\end{align}
with
\begin{align}
    \mu_\theta(x_{t-1}|x_t, t)&=\sqrt{\bar{\alpha}_{t-1}}\hat{x}_0 + \sqrt{1-\bar{\alpha}_{t-1}-\sigma_t^2}\epsilon_\theta(x_t, t)\label{eq:ddpo_mu}\\
    \sigma_t &= \sqrt{\frac{1-\bar{\alpha}_{t-1}}{1-\bar{\alpha}_t}}\sqrt{1-\frac{\bar{\alpha}_t}{\bar{\alpha}_{t-1}}}\label{eq:ddpo_sigma}
\end{align}
as previously proved in Eq.~\eqref{eq:x_t_1_3}, and $\hat{x}_0=\frac{x_t-\sqrt{1-\bar{\alpha}_t}\epsilon_\theta}{\sqrt{\bar{\alpha}_t}}$. Then, the log-probability of Gaussian has explicit form:
\begin{align}
    \log p_\theta(x_{t-1}|x_t) = -\frac{(x_{t-1}-\mu_\theta)^2}{2\sigma_t^2} - \log \sigma_t - \log \sqrt{2\pi}
    \label{eq:ddpo_logprob}
\end{align}
\begin{align}
    \nabla_\theta \log p_\theta(x_{t-1}|x_t) = \frac{\mu_\theta - x_{t-1}}{\sigma_t^2}\nabla_\theta \mu_\theta\nonumber
\end{align}

DDPO requires per-step gradient estimation along the entire diffusion chain. But it does not require reward gradient or high-order sampling gradients along the diffusion chain. 

The method as Eq.~\eqref{eq:ddpo_sf} here is DDPO$_\text{SF}$. This is the online version for gradient estimation, which requires to sample $x_{t-1}$ as well as calculating the probabilities $p_\theta (x_{t-1}|x_t)$ along the sampling process at the same time, such that the model parameters $\theta$ remain the same for sampling and probability estimation. The update only takes one step to preserve the online estimation property. Original paper~\cite{black2023training} also proposes another version for offline policy gradient estimation with importance sampling to allow multi-step updates.
The practical algorithm of DDPO$_\text{SF}$ is shown in Alg.~\ref{alg:ddpo}.

\begin{algorithm}[H]
\caption{DDPO$_\text{SF}$ practical procedure.}
\small
\begin{algorithmic}[1]
\While{\text{train}}
\State \texttt{//Sample from random noise along entire diffusion trajectory}
\State $x_T\leftarrow \varepsilon\sim\mathcal{N}(0, \mathbf{I})$
\For{$t\in [T, \dots, 1]$}
\State Get posterior Gaussian $(\mu_\theta, \sigma)$ with $\epsilon_\theta(x_{t}\text{.detach()},t)$ \quad\texttt{//Eq.~\eqref{eq:ddpo_mu} and \eqref{eq:ddpo_sigma}}
\State Sample $x_{t-1}\sim\mathcal{N}(\mu_\theta, \sigma \mathbf{I})$
\State Estimate $\log p_\theta(x_{t-1}|x_{t})$  \quad\texttt{//Eq.~\eqref{eq:ddpo_logprob}}
\EndFor
\State Get reward $R=R(\hat{x}_0)\text{.detach()}$
\State \texttt{//REINFORCE policy gradient with truncation}
\State $\mathcal{L}_{\text{DDPO}_\text{SF}}=-\sum_n^N \log p_\theta(x_{t_{n-1}}\text{.detach()}|x_{t_n}) \cdot R$ \quad\texttt{//Eq.~\eqref{eq:ddpo_sf}}
\State $\theta\leftarrow \text{GradientDescent}(\theta, \mathcal{L}_{\text{DDPO}_\text{SF}})$ 
\EndWhile
\end{algorithmic}
\label{alg:ddpo}
\end{algorithm}

\paragraph{Diffusion Reward-weighted Regression (DRWR)~\cite{ren2024diffusion}.}
DRWR is a modified version of reward-weighted regression~\cite{peters2007reinforcement} algorithm for reward-based optimization, using a weighted supervised learning objective.

\begin{align}
    \mathcal{L}_\text{DRWR}(\theta)=\mathbb{E}_{x,\varepsilon, t}[\min(e^\beta R(x), w_\text{max})||\varepsilon - \epsilon_\theta(x_t, t)||^2]\nonumber
\end{align}
$\beta$ and $w_\text{max}$ are scaling and clipping coefficients.

To reduce reward estimation variances and improve training stability, the reward value $R(x)$ can be replaced with the advantage estimation (in reinforcement learning terminology) through baseline subtraction: $A(x) = R(x) - R(\bar{x})$, following advantage-weighted regression~\cite{peng2019advantage}. Advanced reinforcement learning techniques like TD-bootstrapped advantage estimation can be further applied as in diffusion advantage-weighted regression (DAWR) method.

\section*{Acknowledgments}
We thank Kexin Jin and Yuheng Zheng for their invaluable assistance in proofreading this draft and providing helpful feedback.




\bibliographystyle{apalike}
\bibliography{refs}





\end{document}